\documentclass[10pt,twocolumn,letterpaper]{article}

\usepackage[pagenumbers]{cvpr} 

\definecolor{cvprblue}{rgb}{0.21,0.49,0.74}
\usepackage[pagebackref,breaklinks,colorlinks,allcolors=cvprblue]{hyperref}
\usepackage{enumitem} 
\usepackage{comment}

\usepackage{booktabs}      
\usepackage{multirow}      
\usepackage{multicol}      
\usepackage{graphicx}      
\usepackage{xcolor}        
\usepackage{colortbl}      
\usepackage{array}         
\usepackage{makecell}      
\usepackage{rotating}      
\usepackage{caption}       
\usepackage{setspace}      

\usepackage{fontawesome5} 
\usepackage[most]{tcolorbox}
\definecolor{BetterGreen}{HTML}{1a7f37}
\definecolor{BetterBlue}{HTML}{1f75cb}
\definecolor{BetterYellow}{HTML}{f5a800}
\definecolor{BetterRed}{HTML}{e00000}

\definecolor{promptgray}{RGB}{235,235,235}

\def\paperID{} 
\def\confName{}
\def\confYear{2026}

\title{Reasoning via Video: The First Evaluation of Video Models’ Reasoning Abilities through Maze-Solving Tasks}

\author{
\textbf{Cheng Yang$^{1}$}$^{*}$ \quad
\textbf{Haiyuan Wan$^{2,3}$}$^{*}$ \quad
\textbf{Yiran Peng$^{1}$}$^{*}$ \quad
\textbf{Xin Cheng$^{4}$} \quad
\textbf{Zhaoyang Yu$^{1}$} \quad
\textbf{Jiayi Zhang$^{1,8}$} \quad\\
\textbf{Junchi Yu$^{5}$}$^{\dagger}$ \quad
\textbf{Xinlei Yu$^{6}$} \quad
\textbf{Xiawu Zheng$^{7}$} \quad
\textbf{Dongzhan Zhou$^{3}$} \quad
\textbf{Chenglin Wu$^{1}$}$^{\dagger}$ \\
$^{1}$DeepWisdom \quad
$^{2}$Tsinghua University \quad
$^{3}$Shanghai Artificial Intelligence Laboratory \\
$^{4}$Renmin University of China \quad
$^{5}$University of Oxford \quad
$^{6}$National University of Singapore \\
$^{7}$Xiamen University \quad
$^{8}$Hong Kong University of Science and Technology (GuangZhou) \\
{\large \faGlobe\;\href{https://imyangc7.github.io/VRBench_Web}{\textcolor{blue}{\textit{https://imyangc7.github.io/VRBench\_Web}}}} \\[4pt]
\vspace{-18pt}
\small $^{*}$ Core contributors \quad\;\; $^{\dagger}$ Corresponding authors
}

\begin{document}

\twocolumn[{%
	\renewcommand\twocolumn[1][]{#1}%
    \maketitle
    \footnotetext[1]{These authors contributed equally to this work.}
    \footnotetext[2]{Corresponding author.}
	\begin{center}
		\vspace{-1em}
		\includegraphics[width=0.95\linewidth]{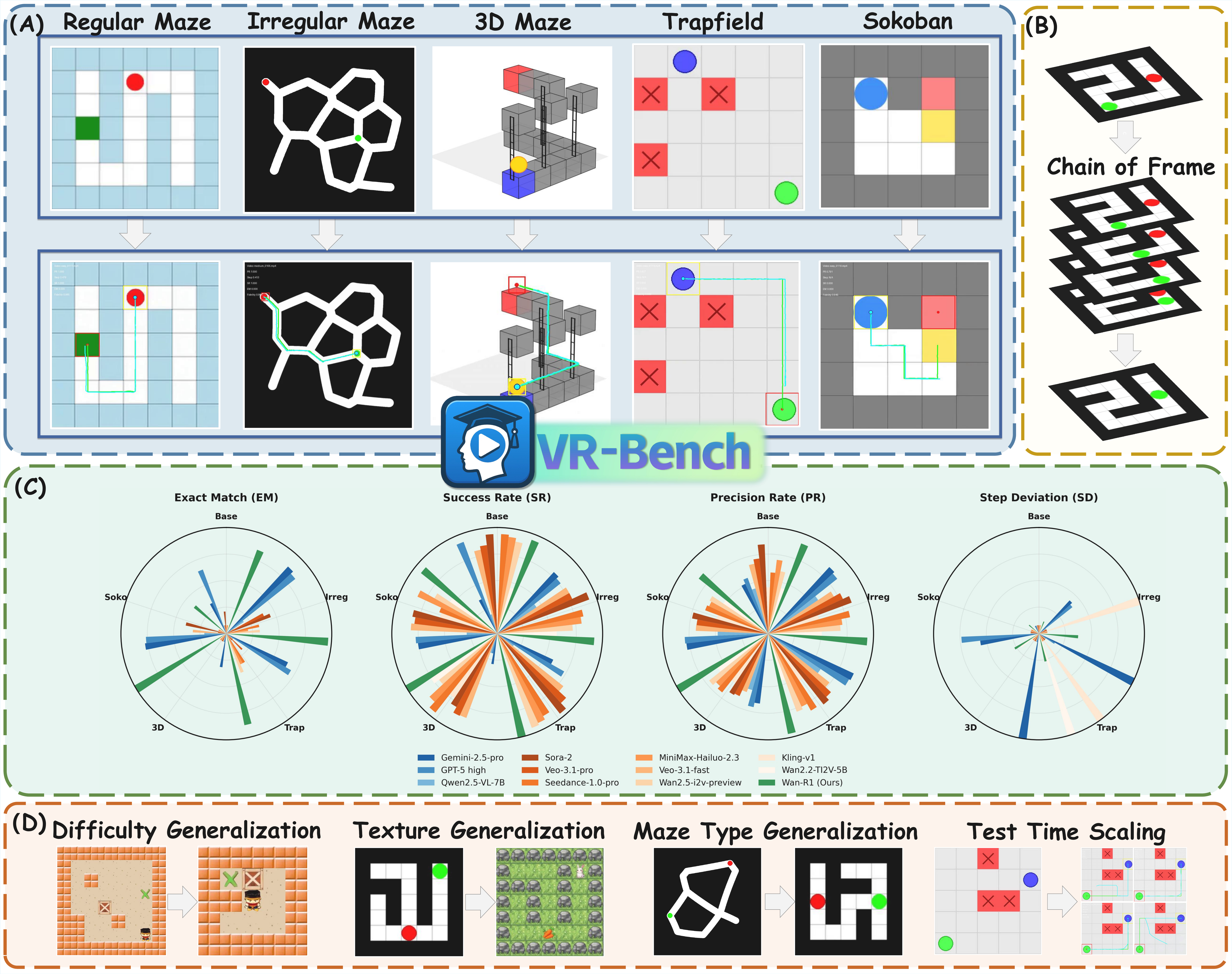}
		\vspace{-0.5em}
		\captionof{figure}{Overview of VR-Bench.
(A) Maze Types. VR-Bench comprises five maze types—Regular Maze, Irregular Maze, 3D Maze, Trapfield, and Sokoban—covering both 2D and 3D settings as well as diverse task structures, yielding a broad range of spatial reasoning scenarios.
(B) Reasoning via Video Paradigm. VR-Bench adopts a chain-of-frame reasoning paradigm \citep{wiedemer2025video}, requiring models to produce frame-by-frame inferences that capture sequential visual reasoning.
(C) Benchmark Performance. Leading VLMs and video models are evaluated on four core metrics across all maze types, revealing clear differences in spatial reasoning capability.
(D) Additional Analysis. VR-Bench also supports evaluations on difficulty generalization, texture generalization, maze-type generalization, and test-time scaling, enabling a comprehensive assessment of model robustness and generalization.}
		\label{fig:teaser}
	\end{center}%
}]

\clearpage
\begin{abstract}
Video Models have achieved remarkable success in high-fidelity video generation with coherent motion dynamics.
Analogous to the development from text generation to text-based reasoning in language modeling, the development of video models motivates us to ask: \textbf{Can video models reason via video generation?}
Compared with the discrete text corpus, video grounds reasoning in explicit spatial layouts and temporal continuity, which serves as an ideal substrate for spatial reasoning.
In this work, we explore the reasoning via video paradigm and introduce VR-Bench—a comprehensive benchmark designed to systematically evaluate video models' reasoning capabilities.
Grounded in maze-solving tasks that inherently require spatial planning and multi-step reasoning, 
VR-Bench contains 7,920 procedurally generated videos across five maze types and diverse visual styles.
Our empirical analysis demonstrates that SFT can efficiently elicit the reasoning ability of video model.
Video models exhibit stronger spatial perception during reasoning, outperforming leading VLMs and generalizing well across diverse scenarios, tasks, and levels of complexity.
We further discover a test-time scaling effect, where diverse sampling during inference improves reasoning reliability by 10–20\%.
These findings highlight the unique potential and scalability of reasoning via video for spatial reasoning tasks.

\end{abstract}    
\vspace{-7mm}
\section{Introduction}
\label{sec:intro}

With the rapid development of diffusion-based and autoregressive-based generative architectures, video models have witnessed tremendous success in high-fidelity video generation.
Previous works, such as Stable Video Diffusion \cite{blattmann2023stable} and Imagen Video \cite{ho2022imagen}, showcase the capability of video models to generate physically realistic and temporally consistent videos conditioned on their input instructions.
Recent studies further reveal that advanced video models are capable of performing a diverse range of visual tasks beyond generation itself, including perception, understanding, and even reasoning.
These findings suggest that video models are evolving from pure generative models into general-purpose visual intelligence models.
Analogous to the evolution of language models from text generation to text-based reasoning, the development of video models leads to a question:
\textbf{“Can video models reason via video generation?”}

Crucially, the spatiotemporal nature of video modality offers a new perspective on reasoning. 
The traditional paradigm, which we term \textit{reasoning via text}, uses language as the medium for expressing intermediate reasoning steps.
Representative work, such as Chain-of-Thought prompting \cite{wei2022chain, zhang2024aflow, yang2025multi, yuthought, wan2025deepresearch, yu2025recode, yao2023tree}, achieves this by eliciting large language models (LLMs) to generate a coherent textual reasoning chain.
Recently, this reasoning via text paradigm has been introduced to visual domains, including multimodal question answering and video understanding.
However, even in these multimodal settings, current paradigms still express reasoning through textual continuation instead of visual or physical dynamics.
In contrast, video represents reasoning as a process of visual continuation over time.
Each frame in a video builds upon its previous ones, capturing the dynamics of motion, spatial consistency, and temporal causality within 2D and 3D space.
The continuous and structured nature of frames makes video an ideal substrate for multimodal reasoning.
Building on this insight, we propose \textit{reasoning via video}, where reasoning emerges through next-frame generation rather than next-token prediction.

However, a comprehensive testbed for reasoning via video is lacking. 
To this end, we introduce VR-Bench, a dedicated benchmark designed to systematically assess the reasoning capabilities of video generation models. 
As shown in Figure~\ref{fig:teaser}, we ground our benchmark in the maze-solving task, a natural fit for visual reasoning due to its open-ended solution space and rich trajectory-based supervision. Each instance inherently demands spatial planning, dynamic tracking, and multi-step reasoning, making it an ideal testbed for evaluating model inference quality over time. Our dataset comprises 7,920 procedurally generated maze-centric videos, each paired with a corresponding Trace Reasoning Task that requires models to infer the optimal path. To ensure broad generalizability and challenge model robustness, VR-Bench spans five distinct maze types—Regular Maze, Irregular Maze, 3D Maze, Sokoban, and Trapfield—covering a wide spectrum of spatial structures and decision patterns. Additionally, each maze is rendered in diverse visual styles across more than a dozen themes, enabling fine-grained analysis of how well models generalize across varied visual domains and increasing the realism and complexity of the reasoning tasks.

Building upon the proposed VR-Bench, we conduct a systematic study of the reasoning via video paradigm.
We construct instruction-following datasets derived from VR-Bench to elicit the reasoning capability of open-source video models.
After supervised fine-tuning (SFT), these models exhibit significant performance gain across all reasoning tasks in VR-Bench.
Moreover, SFT endows video models with strong out-of-domain generalization under diverse distribution shifts, including task difficulty, background style, and task type.
Compared with vision–language models (VLMs) \citep{comanici2025gemini, Qwen2.5-VL, li2024llava, jiang2025mme, chen2025mint} that reason via text, video models consistently outperform their counterparts on high-complexity reasoning tasks, showing greater stability and even superior performance as task difficulty increases, across diverse scenarios and tasks.
This finding confirms that videos serve as a more expressive substrate for spatial reasoning, which facilitates video models to leverage temporal continuity and dynamic visual context.
Interestingly, we further observe that video models exhibit a test-time scaling effect analogous to that of LLMs.
As the inference budget increases, their performance improves substantially.
By employing diverse sampling strategies at test time, video models effectively explore multiple reasoning trajectories, reducing uncertainty and achieving an average performance gain of 10–20\%.
These empirical results highlight the unique potential and scalability of the reasoning via video paradigm.

Our contributions are summarized as follows:
\begin{itemize}
    \item We make an early and systematic exploration of the \textit{reasoning via video} paradigm, where reasoning emerges from sequential frame generation rather than token prediction. Compared with text-based approaches, this paradigm naturally captures temporal continuity and spatial causality, offering a more expressive and scalable substrate for solving spatial reasoning tasks.

    \item We construct \textbf{VR-Bench}, a comprehensive benchmark grounded in maze-solving tasks with diverse spatial structures, difficulty levels, and texture styles. It provides fine-grained trajectory-level supervision and supports evaluations on path accuracy, rule compliance, generalization.

    \item Through extensive experiments, we demonstrate that \textit{video-based reasoning} outperforms \textit{text-based reasoning} (e.g., VLMs) on complex tasks, especially under distribution shifts in maze type, visual style, and difficulty. Fine-tuned video models exhibit stronger performance, lower path redundancy, and higher structural fidelity.

    \item We reveal a test-time scaling effect for video models, where performance consistently improves with larger inference budgets. Similar to that in LLMs, diverse sampling unlocks multi-path exploration and yields up to 20\% performance gains across metrics and difficulty levels.
\end{itemize}

\section{Related Works}
\label{sec:formatting}

\subsection{Video Generation}
Video models have advanced rapidly in both understanding and generation. Early understanding methods, such as MViT \citep{fan2021multiscale}, Video Swin Transformer \citep{liu2022video}, and VideoMAE \citep{tong2022videomae}, focused on learning robust video representations for downstream tasks. With LLMs \cite{blattmann2023stable, ouyang2022training, anil2023palm}, recent approaches tokenize videos and leverage language backbones for captioning~\citep{tong2025g}, event localization~\citep{tian2018audio}, and reasoning~\citep{hu2025video}. On the generation side, Sora-2 \citep{brooks2024video} achieved controllable, physically consistent outputs with synchronized dialogue and sound. Proprietary systems such as Runway’s Gen-3 \citep{runway2024gen3}, Pika Labs \citep{pikalabs2024pika15}, Luma AI \citep{lumalabs2024dreammachine}, and Google DeepMind’s Veo series \citep{deepmind2024veo2, deepmind2025veo3} further enhance video quality and realism but remain closed-source. In contrast, open-source frameworks such as Stable Video Diffusion \citep{blattmann2023stable}, OpenSora\citep{opensora}, Hunyan-Video \citep{kong2024hunyuanvideo}, and the Wan series \citep{wan2025wan} democratize access, offering efficient architectures and scalable training for state-of-the-art video synthesis.

\subsection{Evolution of Reasoning Paradigms}
Chain-of-Thought (CoT) prompting has significantly enhanced the reasoning abilities of language models~\cite{wei2022chain, wang2022self, guo2025deepseek}. Reinforcement learning further integrates CoT-style reasoning into model training, enabling models to internalize multi-step thought processes. More recently, such paradigms have been extended to vision-language models (VLMs). Systems like o3 and o4-mini~\cite{openai2025o3systemcard} introduce the "Think with Image" framework, where reasoning is grounded in visual operations such as zooming and cropping. This allows the model to dynamically interact with image regions as part of the CoT process, thereby improving multimodal reasoning~\cite{zheng2025deepeyes, su2025pixel, zhu2025active}. In parallel, the rise of unified models for both generation and understanding has given birth to a new reasoning paradigm centered on \emph{interleaved vision-language outputs}. Instead of purely textual reasoning traces, these models generate coherent sequences that alternate between textual and visual elements~\cite{xie2025show, wu2025qwen, duan2025got, wan2025deepresearch}, providing a more grounded and expressive format for complex multimodal reasoning.

\subsection{Evaluation of Video Generation Reasoning}

Previous benchmarks for video generation models have predominantly focused on assessing visual quality, temporal coherence, and alignment with human preferences~\cite{huang2024vbench, huang2024vbench++, han2025video, yuan2024chronomagic}. However, these evaluations largely neglect the reasoning capabilities of video models. Recent works have begun to explore \textit{reasoning via video generation}—the ability of models to solve reasoning tasks through the generation process itself~\cite{guo2025video, wiedemer2025video, tong2025thinking}. For instance, models like Veo 3 demonstrate zero-shot competence in tasks such as maze navigation and symmetry recognition. These tasks require perceiving, modeling, and manipulating the visual world, indicating that video generation can inherently support spatial-temporal reasoning. Despite these promising directions, current benchmarks for video reasoning still suffer from several limitations:
\textbf{(1)} \textit{Lack of fine-grained and objective evaluation}: Current evaluations rely heavily on manual inspection or coarse metrics, without capturing the reasoning trajectory embedded in the video; 
\textbf{(2)} \textit{Absence of modality comparisons}: There is a lack of systematic comparison with \textit{think with text} or \textit{think with image} paradigms, making it unclear whether video generation truly provides unique advantages for reasoning; 
\textbf{(3)} \textit{Neglect of tuning and scaling analysis}: Unlike language or multimodal models, video reasoning benchmarks seldom explore whether supervised fine-tuning (SFT) or test-time scaling can improve performance. These gaps call for a new benchmark that evaluates not only generation quality but also the \textit{reasoning process} in videos, using rigorous metrics, multimodal comparisons, and extensible settings.


\vspace{-8pt}
\section{VR-Bench}

\subsection{Dataset Construction}

The VR-Bench dataset is a Visual Trace Reasoning (VTR) dataset that constructs various Maze Puzzles into visual reasoning tasks. Its construction process comprises two steps: Maze Generation and Video Generation.

\begin{figure}[h!]
\centering
\includegraphics[width=1\columnwidth]{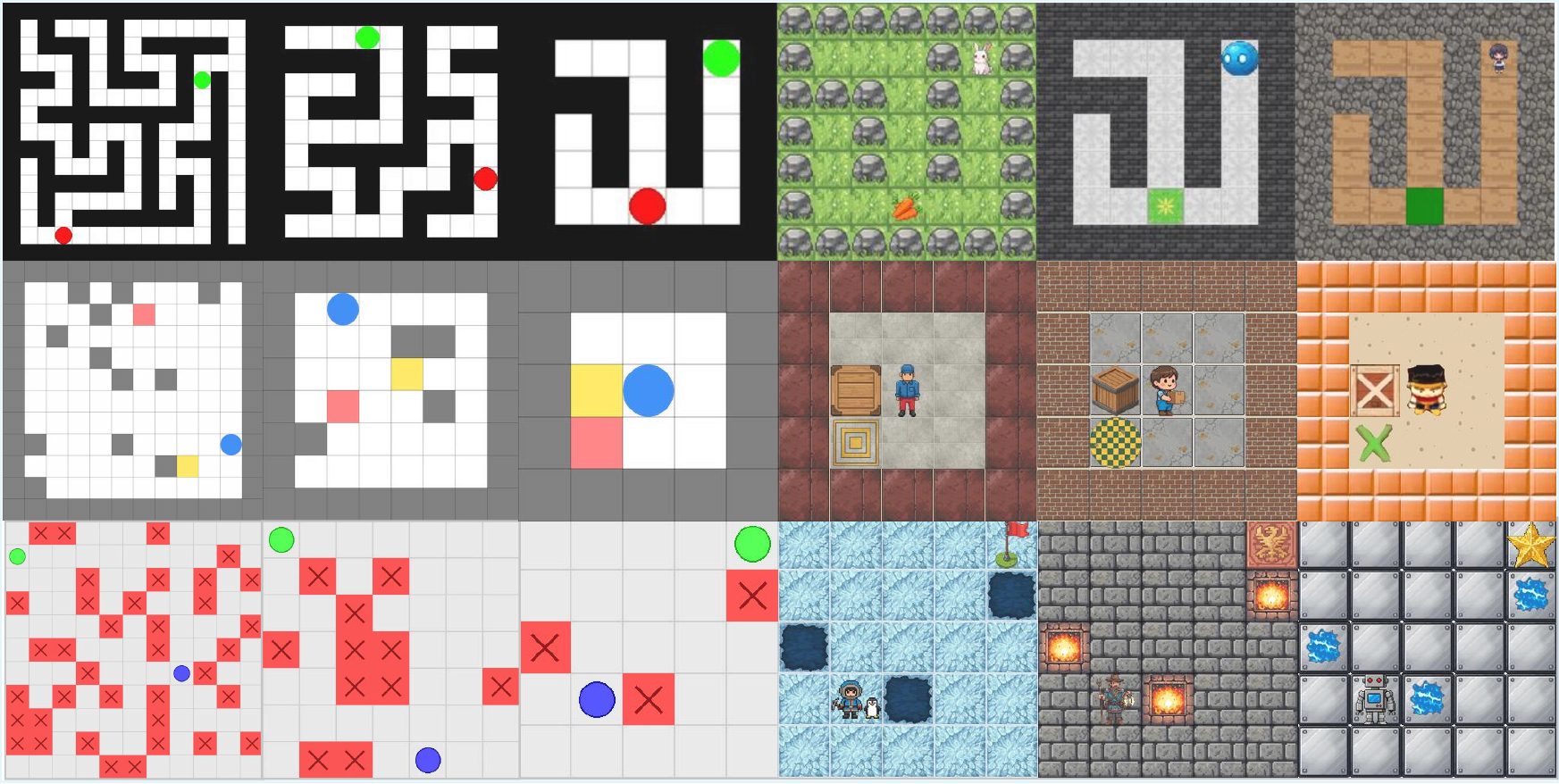} 
\caption{Variations of difficulty level and maze texture}
\label{fig:vary}
\end{figure}
\vspace{-12pt}

\paragraph{Maze Generation.}
The maze generation process in our work encompasses five distinct types, with all 7,920 maze instances in the dataset generated programmatically through custom code \citep{tong2025code2logic}. Each type is tailored to assess specific visual reasoning capabilities, as elaborated below:
\vspace{-2mm}

\begin{table*}[t]
\centering
\setlength{\tabcolsep}{0.9mm}{
\resizebox{0.98\linewidth}{!}{
\begin{tabular}{c|l|ccccc|ccccc|ccccc|ccccc}
\toprule
\multicolumn{2}{c|}{\multirow{2}{*}{\textbf{Method}}} &
\multicolumn{5}{c|}{\textbf{EM}~($\uparrow$)} & 
\multicolumn{5}{c|}{\textbf{SR}~($\uparrow$)} & 
\multicolumn{5}{c|}{\textbf{PR}~($\uparrow$)} & 
\multicolumn{5}{c}{\textbf{SD}~($\downarrow$)} \\
\multicolumn{2}{c|}{} & Base & Irreg & Trap & 3D & Soko & Base & Irreg & Trap & 3D & Soko & Base & Irreg & Trap & 3D & Soko & Base & Irreg & Trap & 3D & Soko \\ \midrule

\multirow{4.5}{*}{\rotatebox{90}{\textbf{VLM}}} 
& Gemini-2.5-pro        & 2.8 & \underline{36.1} & 13.9 & 2.8 & 25.0 & 4.2 & 37.5 & 13.9 & 2.8 & 31.9 & 9.5 & \underline{47.9} & 57.6 & 19.4 & 33.8 & \underline{25.0} & \textbf{1.9} & \textbf{0.0} & \textbf{0.0} & \underline{1.1}  \\
& Gpt-5 high            & \underline{13.9} & 31.9 & 18.1 & 0.0 & \underline{23.6} & 69.4 & 33.3 & 27.8 & 1.4 & 34.7 & 11.8 & 43.6 & 53.7 & 23.7 & 35.6 & 31.0 & \underline{2.1} & 6.9 & 20.0 & \textbf{0.5}  \\
& Qwen2.5-VL-7B$^\heartsuit$         & 0.0 & 1.3 & 0.0 & 0.0 & 1.4 & 1.4 & 6.9 & 1.4 & 1.4 & 2.8 & 6.5 & 12.4 & 14.3 & 11.3 & 7.8 & 300.0 & 26.7 & 66.7 & 80.0 & 1150.0  \\
& Qwen2.5-VL-7B-SFT         & 12.5 & 29.2 & \underline{22.2} & \underline{31.9} & \textbf{29.8} & 52.8 & 34.7 & 52.8 & 36.1 & 37.5 & 32.5 & 45.1 & \underline{71.6} & \underline{59.3} & \underline{43.0} & 52.6 & 2.3 & 11.3 & 4.5 & 3.0  \\ 
& \multicolumn{1}{c|}{$\boldsymbol{\Delta}\!\uparrow$} &
\cellcolor{yellow!20}{+12.5} & \cellcolor{yellow!20}{+27.9} & \cellcolor{yellow!20}{+22.2} & \cellcolor{yellow!20}{+31.9} & \cellcolor{yellow!20}{+28.4} &
\cellcolor{yellow!20}{+51.4} & \cellcolor{yellow!20}{+27.8} & \cellcolor{yellow!20}{+51.4} & \cellcolor{yellow!20}{+34.7} & \cellcolor{yellow!20}{+34.7} &
\cellcolor{yellow!20}{+26.0} & \cellcolor{yellow!20}{+32.7} & \cellcolor{yellow!20}{+57.3} & \cellcolor{yellow!20}{+48.0} & \cellcolor{yellow!20}{+35.2} &
\cellcolor{yellow!20}{-247.4} & \cellcolor{yellow!20}{-24.4} & \cellcolor{yellow!20}{-55.4} & \cellcolor{yellow!20}{-75.5} & \cellcolor{yellow!20}{-1147.0} \\ \midrule 

\multirow{11}{*}{\rotatebox{90}{\textbf{General Video Model}}} 
& \multicolumn{20}{c}{\textbf{Closed-Source}} \\[2pt]
& Veo-3.1-fast          & 0.0 & 0.0 & 0.0 & 0.0 & 2.8 & 40.3 & 36.1 & 38.9 & 48.6 & 43.1 & 20.2 & 24.8 & 28.2 & 13.4 & 21.7 & 195.3 & 111.5 & 80.7 & 33.5 & 112.3 \\ 
& Veo-3.1-pro           & 0.0 & 4.2 & 1.4 & 0.0 & 0.0 & 47.2 & 36.1 & 59.7 & 50.0 & 37.5 & 24.6 & 33.9 & 39.1 & 18.0 & 21.4 & 140.7 & 94.5 & 85.4 & 40.1 & 141.8 \\
& Sora-2                & 1.4 & 5.6 & 0.0 & 0.0 & 4.2 & \underline{75.0} & \textbf{72.2} & \underline{83.0} & 37.5 & 43.1 & \underline{45.1} & 45.7 & 46.6 & 19.3 & 27.4 & 302.9 & 187.0 & 145.1 & 92.4 & 138.7 \\ 
& kling-v1              & 0.0 & 0.0 & 0.0 & 0.0 & 0.0 & 2.8 & 0.0 & 1.4 & 27.8 & 12.5 & 6.3 & 8.8 & 10.4 & 11.7 & 9.0 & 25.2 & -- & -- & 69.7 & 356.1 \\
& Seedance-1.0-pro      & 0.0 & 2.8 & 2.8 & 0.0 & 0.0 & 75.0 & 45.8 & 59.7 & \underline{77.8} & 13.9 & 12.8 & 35.8 & 42.7 & 23.6 & 17.1 & 162.3 & 143.4 & 99.1 & 84.4 & 241.9 \\
& MiniMax-Hailuo-2.3    & 0.0 & 1.4 & 2.8 & 0.0 & 0.0 & 68.1 & 40.3 & 70.8 & 55.6 & \underline{45.8} & 23.2 & 24.2 & 30.3 & 20.3 & 15.5 & 464.0 & 170.0 & 90.9 & 50.1 & 165.5 \\ \cline{2-22} \noalign{\vskip 3pt}

& \multicolumn{20}{c}{\textbf{Open-Source}} \\[2pt]
& Wan2.5-i2v-preview    & 0.0 & 2.8 & 4.2 & 0.0 & 0.0 & 58.3 & 26.4 & 77.8 & 24.5 & 22.4 & 14.3 & 21.8 & 34.4 & 24.5 & 17.1 & 378.4 & 281.8 & 73.2 & 119.9 & 278.0 \\
& Wan2.2-TI2V-5B$^\Diamond$ & 0.0 & 0.0 & 0.0 & 0.0 & 0.0 & 6.9 & 12.5 & 0.0 & 31.9 & 11.1 & 6.6 & 9.1 & 7.1 & 12.8 & 9.2 & 388.7 & 66.1 & -- & 5.4 & 176.6\\ \midrule 

\multirow{1}{*}{\rotatebox{90}{\textbf{Ours}}} 
& \textbf{Wan-R1} & 
\textbf{33.3} & \textbf{56.9} & \textbf{38.9} & \textbf{65.3} & 4.2 & 
\textbf{76.4} & \underline{69.4} & \textbf{100.0} & \textbf{100.0} & \textbf{69.4} & 
\textbf{60.6} & \textbf{71.6} & \textbf{79.1} & \textbf{93.5} & \textbf{44.3} & 
\textbf{10.3} & 2.4 & \underline{3.9} & \underline{3.9} & 10.2 \\
& \multicolumn{1}{c|}{$\boldsymbol{\Delta}\!\uparrow$} &
\cellcolor{yellow!20}{+33.3} & \cellcolor{yellow!20}{+56.9} & \cellcolor{yellow!20}{+38.9} & \cellcolor{yellow!20}{+65.3} & \cellcolor{yellow!20}{+4.2} & 
\cellcolor{yellow!20}{+69.5} & \cellcolor{yellow!20}{+56.9} & \cellcolor{yellow!20}{+100.0} & \cellcolor{yellow!20}{+68.1} & \cellcolor{yellow!20}{+58.3} & 
\cellcolor{yellow!20}{+54.0} & \cellcolor{yellow!20}{+62.5} & \cellcolor{yellow!20}{+72.0} & \cellcolor{yellow!20}{+80.7} & \cellcolor{yellow!20}{+35.1} & 
\cellcolor{yellow!20}{-12.8} & \cellcolor{yellow!20}{-25.1} & \cellcolor{yellow!20}{--} & \cellcolor{yellow!20}{-7.8} & \cellcolor{yellow!20}{-100.1} \\

\bottomrule
\end{tabular}}}
\caption{
The five tasks of VR-Bench correspond to Base (Regular Maze), Irrg (Irregular Maze), Trap (TrapField), 3D (3D Maze), and Soko (Sokoban).
The best and second-best results in each column are \textbf{bolded} and \underline{underlined}, respectively.
``--'' indicates that the model produced no successful cases for the corresponding task, making SD undefined.
$\Diamond$ denotes the base model of Wan-R1, for comparisons.}
\label{main_comparison}
\end{table*}

\begin{enumerate}[label={\bfseries \arabic*.}, leftmargin=*, topsep=6pt, itemsep=3pt]
    \item \textbf{Regular Maze.}
    We generate mazes with a grid-based layout to focus on the model's ability to perceive basic maze structures and its path-finding and problem-solving competence, serving as a fundamental testbed for maze reasoning tasks.

    \item \textbf{Trapfield.}
    This type transforms the "walls" of traditional mazes into grid-shaped trap regions, reversing the logic from "finding feasible paths" to "avoiding traps". Beyond altering the problem-solving logic, the more flexible movement space also challenges the model's ability to plan and find optimal paths.

    \item \textbf{Irregular Maze.}
    Moving away from regular block-shaped paths, we adopt curve-based path designs. This design prevents the model from relying on coordinate-based position encoding, thereby rigorously evaluating its pure visual perception of maze layouts. It also explicitly decouples visual reasoning from text-based reasoning, focusing on the video model's capability to reason via video itself.

    \item \textbf{Sokoban.}
    We modify the underlying rules of traditional mazes by introducing the "Sokoban" task mechanism. Models need to comprehend and apply Sokoban logic on top of path finding, increasing task complexity and emphasizing the model's ability to internalize and apply logic.

    \item \textbf{3D Maze.}
    By extending the maze to a 3D space, we employ a stereoscopic structural design to test the model's spatial perception ability in 3D environments and its cross-dimensional path reasoning capability.
\end{enumerate}
\vspace{-6mm}

\paragraph{Maze Variations.}
To evaluate the generalization ability on the VTR task and enhance robustness in adapting to diverse maze scenarios, we introduce variations across two key dimensions: (1) \textbf{Difficulty Level}: We define three difficulty grades (Easy, Medium, and Hard) by adjusting the maze size (e.g., expanding from 5×5 to 7×7), modifying the number of maze branches, and adding obstacles; (2) \textbf{Maze Texture}: We vary the textures of maze obstacles, paths, and other components using textures generated via procedural methods and generative models, as shown in Figure \ref{fig:vary}, which exposes the policies to a broad visual distribution and mitigates overfitting to clean, synthetic environments.

\vspace{-10pt}
\paragraph{Video Generation.}

To generate solution videos from maze images, we use a Breadth-First Search solver to compute the optimal path for each maze. These paths are rendered into videos at 24 fps and standardized to 192 frames (8 seconds) by adjusting playback speed, producing consistent image–video pairs for training and evaluation.

\vspace{-4pt}
\subsection{Metric Design}
We selected two different evaluation paradigms to comprehensively assess our task, as detailed below.
\vspace{-4mm}
\paragraph{Path Matching.} 
To objectively and comprehensively evaluate the VTR task, we perform target tracking across each frame of the model-generated videos to record the motion trajectory of the target. By comparing and analyzing these trajectories against the optimal path for each task \citep{xu2025visual}, we propose the following four evaluation metrics.

\begin{enumerate}[label={\bfseries \arabic*.}, leftmargin=*, topsep=6pt, itemsep=3pt]
    \item \textbf{Exact Match (EM)} \\
    Defined as $\mathrm{EM}_i = \prod_{j=1}^{n_i} \mathbb{I}(\hat{v}_{ij} = v_{ij})$. This metric measures whether the model successfully generates the complete and correct trajectory that aligns with the shortest optimal valid path. One step of deviation from the optimal solution is considered incorrect.
    
    \item \textbf{Success Rate (SR)} \\
    Defined as $\mathrm{SR}_i = \mathbb{I}\!\left(\mathbf{p}^{(gen)}_{\text{end}}\in\mathcal{B}_{\text{goal}}\right)$. SR measures whether the generated trajectory successfully reaches the designated goal region. It reflects the model’s capability to complete the task by arriving at the target position, with a value of $1$ indicating successful goal attainment and $0$ indicating failure to reach the goal.
    
    \item \textbf{Precision Rate (PR)} \\
    Defined as $\mathrm{PR}_i = \frac{1}{n_i} \sum_{j=1}^{n_i} \left[\prod_{k=1}^{j} \mathbb{I}(\hat{v}_{ik} = v_{ik})\right]$.
    PR quantifies the proportion of consecutively correct steps along the optimal path. 
    It offers a softer metric than EM, reflecting the model’s ability to make steady, meaningful progress toward the complete correct trajectory.

    \item \textbf{Step Deviation (SD)} \\
    Defined as $\mathrm{SD}_i = \frac{L^{(gen)}_i}{L^{(gt)}_i} - 1$. SD quantifies the relative path-length redundancy of the generated trajectory, representing how much longer the model’s path is compared to the optimal one. A smaller SD indicates higher efficiency and closer adherence to the optimal solution.
\end{enumerate}

\begin{figure}[h!]
\centering
\includegraphics[width=0.9\columnwidth]{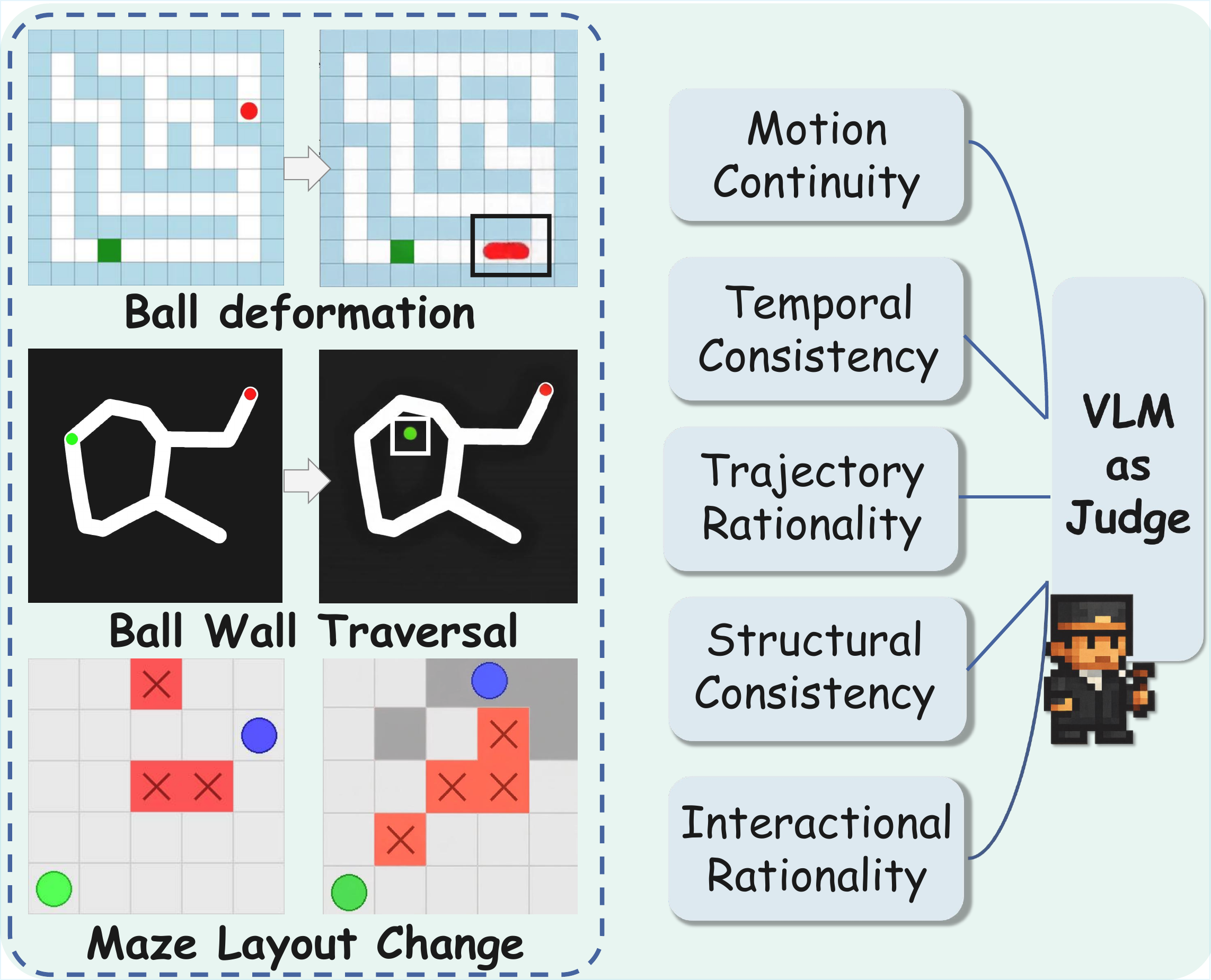} 
\caption{Bad case visualization and VLM-as-judge schematic}
\label{fig:judge}
\end{figure}
\vspace{-18pt}

\paragraph{Rule Compliance.} 
Not all generated videos faithfully depict the ball following the maze path to complete its task. As shown in Figure~\ref{fig:vary}, we observed numerous failure cases during testing, including the ball “breaking through maze walls,” “disappearing and reappearing,” and “inconsistent maze layouts across frames.” To consistently evaluate the model’s adherence to spatial and physical rules \citep{huang2024vbench++}, we designed a prompt-based assessment protocol that examines five key dimensions: (1) \textbf{Motion Continuity} of the main subject, (2) \textbf{Temporal Consistency} of the subject, (3) \textbf{Trajectory Rationality} of the main subject, (4) \textbf{Structural Consistency} of the maze, and (5) \textbf{Interactional Rationality} of subject–maze interactions. Each generated video is interpreted by a VLM, which identifies and scores potential violations of these rules. Specifically, each dimension is assigned a binary score—0 if the behavior is deemed unreasonable and 1 if it is reasonable. The scores from all five dimensions are then aggregated to compute a unified \textbf{VLM-score}, ranging from 0 to 5, which provides a quantitative measure of the overall rule compliance in the generated videos.

Since Structural Consistency offers only a binary judgment of whether the maze layout has changed, we introduce \textbf{Maze Fidelity (MF)} to quantitatively measure the degree of structural consistency across frames, which defined as $\mathrm{MF} = \frac{1}{M} \sum_{i=1}^{M} \left(1 - \frac{|{p : |I_0(p) - I_i(p)| > \tau}|}{N_i}\right)$.

Here, $M$ is the number of sampled frames; $I_0$ and $I_i$ denote the background regions of the first and the $i$-th frames; $\tau$ is the pixel-difference threshold; and $N_i$ is the number of valid overlapping pixels. MF quantifies background stability across frames, with higher values indicating better preservation of the static maze layout.

\vspace{-10pt}
\section{Experiment}
\label{sec:formatting}

\begin{table}[t]
\centering
\setlength{\tabcolsep}{0.9mm}{
\resizebox{0.48\textwidth}{!}{
\begin{tabular}{c|l|ccccc|ccccc}
\toprule
\multicolumn{2}{c|}{\multirow{2}{*}{\textbf{Method}}} &
\multicolumn{5}{c|}{\textbf{MF}~($\uparrow$)} & 
\multicolumn{5}{c}{\textbf{VLM-Score}~($\uparrow$)} \\
& & Base & Irreg & Trap & 3D & Soko 
  & Base & Irreg & Trap & 3D & Soko \\ \midrule

\multirow{11}{*}{\rotatebox{90}{\textbf{General Video Model}}} & \multicolumn{1}{c}{\textbf{Closed-Source}} \\
& Veo-3.1-fast          & 43.2 & 86.3 & 22.5 & 69.1 & 63.5 & 0.8 & 2.5 & 1.1 & 1.7 & 2.0 \\ 
& Veo-3.1-pro           & 80.5 & 89.3 & 82.2 & 73.4 & \textbf{95.8} & 2.5 & 2.8 & 1.5 & 2.0 & 2.5 \\
& Sora-2                & \textbf{96.5} & 97.2 & \textbf{97.2} & \underline{95.4} & \underline{95.2} & \underline{3.9} & \underline{4.1} & \underline{3.9} & \underline{3.3} & 3.2 \\ 
& kling-v1              & 55.5 & 73.8 & 54.4 & 84.9 & 72.9 & 1.4 & 2.5 & 1.9 & 2.8 & 2.8 \\
& Seedance-1.0-pro      & 87.7 & 97.2 & 64.3 & 86.3 & 83.0 & 3.0 & 3.8 & 2.4 & 2.7 & 3.2 \\
& MiniMax-Hailuo-2.3    & \underline{92.4} & 93.4 & 91.4 & 94.9 & 93.3 & 3.5 & 3.4 & 3.2 & 3.7 & \underline{3.6} \\ \cline{2-12} \noalign{\vskip 3pt}

& \multicolumn{1}{c}{\textbf{Open-Source}} & \\
& Wan2.5-i2v-preview    & 69.2 & 74.8 & 70.6 & 82.7 & 90.4 & 1.2 & 1.8 & 2.6 & 2.6 &  2.9\\
& Wan2.2-TI2V-5B$^\Diamond$ & 85.5 &\underline{97.4} & 83.7 & 94.7 & 93.5 & 2.8 & 3.1 & 1.5 & 3.3 & 3.1 \\ \midrule 

\multirow{1}{*}{\rotatebox{90}{\textbf{Ours}}} 
& \textbf{Wan-R1}                       & 91.2 & \textbf{98.1} & \underline{93.3} & \textbf{95.7} & 94.1 & \textbf{4.2} & \textbf{4.3} & \textbf{4.4} & \textbf{4.0} & \textbf{4.1} \\
& \multicolumn{1}{l|}{\textbf{-Wan2.2-TI2V-5B}$^\Diamond$} & \cellcolor{yellow!20}{+5.7} & \cellcolor{yellow!20}{+0.7} & \cellcolor{yellow!20}{+9.6} & \cellcolor{yellow!20}{+1.0} & \cellcolor{yellow!20}{+0.6} & \cellcolor{yellow!20}{+0.3} & \cellcolor{yellow!20}{+0.2} & \cellcolor{yellow!20}{+0.5} & \cellcolor{yellow!20}{+0.7} & \cellcolor{yellow!20}{+0.5} \\
\bottomrule
\end{tabular}}}
\caption{MF and VLM-Score denote Maze Fidelity and the rule-compliance score evaluated by a VLM. The best and second-best results in each column are \textbf{bolded} and \underline{underlined}.}
\label{comparison}
\end{table}

To comprehensively evaluate the reasoning capability of video models, we conduct experiments on our proposed VR-Bench. We evaluate both state-of-the-art proprietary and open-source video models on this benchmark. To highlight the advantages of video models over traditional multimodal approaches, we also include representative VLMs in our evaluation. In addition, we fine-tune the open-source video model \textbf{Wan2.2-TI2V-5B} using VR-Bench to investigate their generalization ability on reasoning tasks. This allows us to assess whether such reasoning capabilities can emerge via supervised fine-tuning, and whether they can generalize across different settings.

\subsection{Training Configurations}

To investigate how well open-source video models can acquire and generalize reasoning abilities through fine-tuning, we trained \textbf{Wan-R1} based on the proposed dataset.

Specifically, we used the first scene from each of the five game types in our benchmark. For each game, we created two training settings: one using only easy samples, and another using a mixture of easy, medium, and hard samples. In each case, the data was split into 80\% for training and 20\% for validation.

All models were fine-tuned using the Accelerate framework on A100 GPUs. We adopted a LoRA-based training strategy on the Wan2.2-TI2V-5B architecture, with a learning rate of 1e-4, image resolution of 512$\times$512, and video length of 193 frames. We applied LoRA (rank 32) to key attention and feedforward modules (q, k, v, o, ffn.0, ffn.2) of the Dit backbone. Each model was trained for 5 epochs, with a dataset repetition factor of 100. Other training parameters such as batch size and GPU days are in Appendix.

\subsection{Baseline Model}

We compare the Wan-R1 against a wide range of baselines: \textbf{\textit{1)}} 6 Closed-source video models: \textit{Veo-3.1-fast}, \textit{Veo-3.1-pro} \citep{deepmind2025veo3}, \textit{Sora-2} \citep{openai2025sora2}, \textit{Kling-v1} \citep{KlingAI2025}, \textit{Seedance-1.0-pro} \citep{Seedance10Pro}, \textit{MiniMax-Hailuo-2.3} \citep{MiniMaxHailuo23}.
\textbf{\textit{2)}} 2 Open-source video models: \textit{Wan2.5-i2v-preview} \citep{AlibabaWan25}, \textit{Wan2.2-TI2V-5B} \citep{wan2025wan}.
\textbf{\textit{3)}} 3 VLMs: \textit{Gemini-2.5-pro} \citep{comanici2025gemini}, \textit{Gpt-5 high} \citep{GPT5ChatGPT}, \textit{Qwen2.5-VL-7B} \citep{Qwen2.5-VL}. The settings of the baseline are in the appendix.

\vspace{-6pt}
\section{Insights and Discussions}

\begin{table}[t]
\centering
\resizebox{\linewidth}{!}{
\begin{tabular}{l|ccc|ccc|ccc|ccc}
\toprule
\textbf{Task} 
& \multicolumn{3}{c|}{\textbf{EM}} 
& \multicolumn{3}{c|}{\textbf{SR}} 
& \multicolumn{3}{c|}{\textbf{PR}} 
& \multicolumn{3}{c}{\textbf{SD}} \\
& Easy & Med. & Hard & Easy & Med. & Hard & Easy & Med. & Hard & Easy & Med. & Hard \\
\midrule

\multirow{2}{*}{Base} 
& 0.0 & 0.0 & 0.0 
& 8.3 & 8.3 & 4.2
& 13.2 & 3.8 & 2.7
& 154.8 & -- & -- 
\\
& 
\makecell[c]{0.0 \\ \textcolor{gray}{\scriptsize (+0.0)}}
& \makecell[c]{4.2 \\ \textcolor{BetterGreen}{\scriptsize (+4.2)}}
& \makecell[c]{0.0 \\ \textcolor{gray}{\scriptsize (+0.0)}}

& \makecell[c]{83.3 \\ \textcolor{BetterGreen}{\scriptsize (+75.0)}}
& \makecell[c]{41.7 \\ \textcolor{BetterGreen}{\scriptsize (+33.4)}}
& \makecell[c]{58.3 \\ \textcolor{BetterGreen}{\scriptsize (+54.1)}}

& \makecell[c]{40.1 \\ \textcolor{BetterGreen}{\scriptsize (+26.9)}}
& \makecell[c]{26.1 \\ \textcolor{BetterGreen}{\scriptsize (+22.3)}}
& \makecell[c]{6.0 \\ \textcolor{BetterGreen}{\scriptsize (+3.3)}}

& \makecell[c]{28.2 \\ \textcolor{BetterGreen}{\scriptsize (-126.6)}}
& \makecell[c]{10.1 \\ \textcolor{gray}{\scriptsize (--)}}
& \makecell[c]{-- \\ \textcolor{gray}{\scriptsize (--)}}
\\
\midrule

\multirow{2}{*}{Irrg}
& 0.0 & 0.0 & 0.0
& 29.2 & 4.2 & 4.2
& 13.4 & 8.1 & 5.8
& 48.3 & -- & 39.3 
\\
&
\makecell[c]{83.3 \\ \textcolor{BetterGreen}{\scriptsize (+83.3)}}
& \makecell[c]{66.7 \\ \textcolor{BetterGreen}{\scriptsize (+66.7)}}
& \makecell[c]{54.2 \\ \textcolor{BetterGreen}{\scriptsize (+54.2)}}

& \makecell[c]{95.8 \\ \textcolor{BetterGreen}{\scriptsize (+66.6)}}
& \makecell[c]{87.5 \\ \textcolor{BetterGreen}{\scriptsize (+83.3)}}
& \makecell[c]{62.5 \\ \textcolor{BetterGreen}{\scriptsize (+58.3)}}

& \makecell[c]{88.0 \\ \textcolor{BetterGreen}{\scriptsize (+74.6)}}
& \makecell[c]{74.8 \\ \textcolor{BetterGreen}{\scriptsize (+66.7)}}
& \makecell[c]{68.4 \\ \textcolor{BetterGreen}{\scriptsize (+62.6)}}

& \makecell[c]{3.5 \\ \textcolor{BetterGreen}{\scriptsize (-44.8)}}
& \makecell[c]{7.8 \\ \textcolor{gray}{\scriptsize (--)}}
& \makecell[c]{3.1 \\ \textcolor{BetterGreen}{\scriptsize (-36.2)}}
\\
\midrule

\multirow{2}{*}{Trap}
& 0.0 & 0.0 & 0.0
& 0.0 & 0.0 & 0.0
& 6.0 & 6.4 & 8.9
& -- & -- & --
\\

&
\makecell[c]{62.5 \\ \textcolor{BetterGreen}{\scriptsize (+62.5)}}
& \makecell[c]{0.0 \\ \textcolor{gray}{\scriptsize (+0.0)}}
& \makecell[c]{12.5 \\ \textcolor{BetterGreen}{\scriptsize (+12.5)}}

& \makecell[c]{100.0 \\ \textcolor{BetterGreen}{\scriptsize (+100.0)}}
& \makecell[c]{95.8 \\ \textcolor{BetterGreen}{\scriptsize (+95.8)}}
& \makecell[c]{62.5 \\ \textcolor{BetterGreen}{\scriptsize (+62.5)}}

& \makecell[c]{86.7 \\ \textcolor{BetterGreen}{\scriptsize (+80.7)}}
& \makecell[c]{43.7 \\ \textcolor{BetterGreen}{\scriptsize (+37.3)}}
& \makecell[c]{56.7 \\ \textcolor{BetterGreen}{\scriptsize (+47.8)}}

& \makecell[c]{2.1 \\ \textcolor{gray}{\scriptsize (--)}}
& \makecell[c]{5.9 \\ \textcolor{gray}{\scriptsize (--)}}
& \makecell[c]{3.5 \\ \textcolor{gray}{\scriptsize (--)}}
\\
\midrule

\multirow{2}{*}{3D}
& 0.0 & 0.0 & 0.0
& 54.2 & 16.7 & 25.0
& 7.6 & 9.8 & 15.2
& 53.0 & 87.9 & 33.6
\\

&
\makecell[c]{41.7 \\ \textcolor{BetterGreen}{\scriptsize (+41.7)}}
& \makecell[c]{0.0 \\ \textcolor{gray}{\scriptsize (+0.0)}}
& \makecell[c]{4.2 \\ \textcolor{BetterGreen}{\scriptsize (+4.2)}}

& \makecell[c]{100.0 \\ \textcolor{BetterGreen}{\scriptsize (+45.8)}}
& \makecell[c]{79.2 \\ \textcolor{BetterGreen}{\scriptsize (+62.5)}}
& \makecell[c]{83.3 \\ \textcolor{BetterGreen}{\scriptsize (+58.3)}}

& \makecell[c]{78.7 \\ \textcolor{BetterGreen}{\scriptsize (+71.1)}}
& \makecell[c]{47.4 \\ \textcolor{BetterGreen}{\scriptsize (+37.6)}}
& \makecell[c]{58.7 \\ \textcolor{BetterGreen}{\scriptsize (+43.5)}}

& \makecell[c]{4.6 \\ \textcolor{BetterGreen}{\scriptsize (-48.4)}}
& \makecell[c]{10.2 \\ \textcolor{BetterGreen}{\scriptsize (-77.7)}}
& \makecell[c]{13.8 \\ \textcolor{BetterGreen}{\scriptsize (-19.8)}}
\\
\midrule

\multirow{2}{*}{Soko}
& 0.0 & 0.0 & 0.0
& 20.8 & 8.3 & 4.2
& 14.6 & 8.3 & 5.6
& 354.2 & 61.4 & --
\\

&
\makecell[c]{4.2 \\ \textcolor{BetterGreen}{\scriptsize (+4.2)}}
& \makecell[c]{0.0 \\ \textcolor{gray}{\scriptsize (+0.0)}}
& \makecell[c]{0.0 \\ \textcolor{gray}{\scriptsize (+0.0)}}

& \makecell[c]{83.3 \\ \textcolor{BetterGreen}{\scriptsize (+62.5)}}
& \makecell[c]{45.8 \\ \textcolor{BetterGreen}{\scriptsize (+37.5)}}
& \makecell[c]{33.3 \\ \textcolor{BetterGreen}{\scriptsize (+29.1)}}

& \makecell[c]{62.2 \\ \textcolor{BetterGreen}{\scriptsize (+47.6)}}
& \makecell[c]{12.5 \\ \textcolor{BetterGreen}{\scriptsize (+4.2)}}
& \makecell[c]{10.4 \\ \textcolor{BetterGreen}{\scriptsize (+4.8)}}

& \makecell[c]{18.8 \\ \textcolor{BetterGreen}{\scriptsize (-335.4)}}
& \makecell[c]{84.5 \\ \textcolor{red}{\scriptsize (+23.1)}}
& \makecell[c]{16.1 \\ \textcolor{gray}{\scriptsize (--)}}
\\

\bottomrule
\end{tabular}}
\caption{
Difficulty generalization of Wan-R1 on VR-Bench. 
Each task block compares the baseline (Wan2.2-TI2V-5B) and the fine-tuned model (trained only on Easy level) across difficulty levels (Easy, Medium, Hard) and four metrics (EM, SR, PR, SD). 
\textcolor{BetterGreen}{Green} indicates improvements, 
\textcolor{red}{red} indicates degradation, 
\textcolor{gray}{gray} denotes no change or undefined cases.
}
\label{tab:difficulty_single}
\end{table}

\begin{table*}[t]
\centering
\resizebox{\textwidth}{!}{
\begin{tabular}{ll|ccccc|ccccc|ccccc|ccccc}
\toprule
\textbf{Task} & \textbf{Model} 
& \multicolumn{5}{c|}{\textbf{EM}} 
& \multicolumn{5}{c|}{\textbf{SR}} 
& \multicolumn{5}{c|}{\textbf{PR}} 
& \multicolumn{5}{c}{\textbf{SD}} \\
& & Base & Irreg & Trap & 3D & Soko 
  & Base & Irreg & Trap & 3D & Soko 
  & Base & Irreg & Trap & 3D & Soko 
  & Base & Irreg & Trap & 3D & Soko \\
\midrule

\multicolumn{2}{l|}{\textbf{Wan2.2-TI2V-5B Baseline}}  
& 0.0 & 0.0 & 0.0 & 0.0 & 0.0   
& 6.9 & 12.5 & 0.0 & 31.9 & 11.1   
& 6.6 & 9.1 & 7.1 & 12.8 & 9.2   
& 388.7 & 66.1 & -- & 5.4 & 176.6 \\
\midrule

\multirow{1}{*}{Regular Maze} 
& Fine-tuned 
& \makecell[c]{33.3 \\ \textcolor{BetterGreen}{\scriptsize (+33.3)}} 
& \makecell[c]{5.6 \\ \textcolor{BetterGreen}{\scriptsize (+5.6)}} 
& \makecell[c]{1.4 \\ \textcolor{BetterGreen}{\scriptsize (+1.4)}} 
& \makecell[c]{0.0 \\ \textcolor{gray}{\scriptsize (+0.0)}} 
& \makecell[c]{0.0 \\ \textcolor{gray}{\scriptsize (+0.0)}} 

& \makecell[c]{76.4 \\ \textcolor{BetterGreen}{\scriptsize (+69.5)}} 
& \makecell[c]{8.3 \\ \textcolor{red}{\scriptsize (-4.2)}} 
& \makecell[c]{88.9 \\ \textcolor{BetterGreen}{\scriptsize (+88.9)}} 
& \makecell[c]{69.4 \\ \textcolor{BetterGreen}{\scriptsize (+37.5)}} 
& \makecell[c]{30.6 \\ \textcolor{BetterGreen}{\scriptsize (+19.5)}} 

& \makecell[c]{60.6 \\ \textcolor{BetterGreen}{\scriptsize (+54.0)}} 
& \makecell[c]{22.7 \\ \textcolor{BetterGreen}{\scriptsize (+13.6)}} 
& \makecell[c]{25.2 \\ \textcolor{BetterGreen}{\scriptsize (+18.1)}} 
& \makecell[c]{13.7 \\ \textcolor{BetterGreen}{\scriptsize (+0.9)}} 
& \makecell[c]{19.0 \\ \textcolor{BetterGreen}{\scriptsize (+9.8)}} 

& \makecell[c]{10.3 \\ \textcolor{BetterGreen}{\scriptsize (-378.4)}} 
& \makecell[c]{51.7 \\ \textcolor{BetterGreen}{\scriptsize (-14.4)}} 
& \makecell[c]{3.8 \\ \textcolor{gray}{\scriptsize (--)}}
& \makecell[c]{12.7 \\ \textcolor{red}{\scriptsize (+7.3)}} 
& \makecell[c]{49.3 \\ \textcolor{BetterGreen}{\scriptsize (-127.3)}} 
\\
\midrule

\multirow{1}{*}{Irregular Maze} 
& Fine-tuned 
& \makecell[c]{0.0 \\ \textcolor{gray}{\scriptsize (+0.0)}} 
& \makecell[c]{56.9 \\ \textcolor{BetterGreen}{\scriptsize (+56.9)}} 
& \makecell[c]{0.0 \\ \textcolor{gray}{\scriptsize (+0.0)}} 
& \makecell[c]{0.0 \\ \textcolor{gray}{\scriptsize (+0.0)}} 
& \makecell[c]{0.0 \\ \textcolor{gray}{\scriptsize (+0.0)}} 

& \makecell[c]{11.1 \\ \textcolor{BetterGreen}{\scriptsize (+4.2)}} 
& \makecell[c]{69.4 \\ \textcolor{BetterGreen}{\scriptsize (+56.9)}} 
& \makecell[c]{52.8 \\ \textcolor{BetterGreen}{\scriptsize (+52.8)}} 
& \makecell[c]{79.2 \\ \textcolor{BetterGreen}{\scriptsize (+47.3)}} 
& \makecell[c]{12.5 \\ \textcolor{BetterGreen}{\scriptsize (+1.4)}} 

& \makecell[c]{16.6 \\ \textcolor{BetterGreen}{\scriptsize (+10.0)}} 
& \makecell[c]{71.6 \\ \textcolor{BetterGreen}{\scriptsize (+62.5)}} 
& \makecell[c]{18.1 \\ \textcolor{BetterGreen}{\scriptsize (+11.0)}} 
& \makecell[c]{16.8 \\ \textcolor{BetterGreen}{\scriptsize (+4.0)}} 
& \makecell[c]{15.5 \\ \textcolor{BetterGreen}{\scriptsize (+6.3)}} 

& \makecell[c]{35.8 \\ \textcolor{BetterGreen}{\scriptsize (-352.9)}} 
& \makecell[c]{2.4 \\ \textcolor{BetterGreen}{\scriptsize (-63.7)}} 
& \makecell[c]{6.9 \\ \textcolor{gray}{\scriptsize (--)}}
& \makecell[c]{9.0 \\ \textcolor{red}{\scriptsize (+3.6)}} 
& \makecell[c]{40.7 \\ \textcolor{BetterGreen}{\scriptsize (-135.9)}} 
\\
\midrule

\multirow{1}{*}{3D Maze} 
& Fine-tuned 
& \makecell[c]{0.0 \\ \textcolor{gray}{\scriptsize (+0.0)}} 
& \makecell[c]{5.6 \\ \textcolor{BetterGreen}{\scriptsize (+5.6)}} 
& \makecell[c]{0.0 \\ \textcolor{gray}{\scriptsize (+0.0)}} 
& \makecell[c]{65.3 \\ \textcolor{BetterGreen}{\scriptsize (+65.3)}} 
& \makecell[c]{0.0 \\ \textcolor{gray}{\scriptsize (+0.0)}} 

& \makecell[c]{38.9 \\ \textcolor{BetterGreen}{\scriptsize (+32.0)}} 
& \makecell[c]{31.9 \\ \textcolor{BetterGreen}{\scriptsize (+19.4)}} 
& \makecell[c]{30.6 \\ \textcolor{BetterGreen}{\scriptsize (+30.6)}} 
& \makecell[c]{100.0 \\ \textcolor{BetterGreen}{\scriptsize (+68.1)}} 
& \makecell[c]{20.8 \\ \textcolor{BetterGreen}{\scriptsize (+9.7)}} 

& \makecell[c]{6.8 \\ \textcolor{BetterGreen}{\scriptsize (+0.2)}} 
& \makecell[c]{20.6 \\ \textcolor{BetterGreen}{\scriptsize (+11.5)}} 
& \makecell[c]{6.7 \\ \textcolor{BetterGreen}{\scriptsize (-0.4)}} 
& \makecell[c]{93.5 \\ \textcolor{BetterGreen}{\scriptsize (+80.7)}} 
& \makecell[c]{15.0 \\ \textcolor{BetterGreen}{\scriptsize (+5.8)}} 

& \makecell[c]{108.2 \\ \textcolor{BetterGreen}{\scriptsize (-280.5)}} 
& \makecell[c]{10.9 \\ \textcolor{BetterGreen}{\scriptsize (-55.2)}} 
& \makecell[c]{8.8 \\ \textcolor{gray}{\scriptsize (--)}}
& \makecell[c]{3.9 \\ \textcolor{BetterGreen}{\scriptsize (-1.5)}} 
& \makecell[c]{80.6 \\ \textcolor{BetterGreen}{\scriptsize (-96.0)}} 
\\
\midrule

\multirow{1}{*}{Sokoban} 
& Fine-tuned 
& \makecell[c]{0.0 \\ \textcolor{gray}{\scriptsize (+0.0)}} 
& \makecell[c]{1.4 \\ \textcolor{BetterGreen}{\scriptsize (+1.4)}} 
& \makecell[c]{1.4 \\ \textcolor{BetterGreen}{\scriptsize (+1.4)}} 
& \makecell[c]{0.0 \\ \textcolor{gray}{\scriptsize (+0.0)}} 
& \makecell[c]{4.2 \\ \textcolor{BetterGreen}{\scriptsize (+4.2)}} 

& \makecell[c]{0.5 \\ \textcolor{red}{\scriptsize (-6.4)}} 
& \makecell[c]{22.2 \\ \textcolor{BetterGreen}{\scriptsize (+9.7)}} 
& \makecell[c]{18.1 \\ \textcolor{BetterGreen}{\scriptsize (+18.1)}} 
& \makecell[c]{22.2 \\ \textcolor{red}{\scriptsize (-9.7)}} 
& \makecell[c]{69.4 \\ \textcolor{BetterGreen}{\scriptsize (+58.3)}} 

& \makecell[c]{15.7 \\ \textcolor{BetterGreen}{\scriptsize (+9.1)}} 
& \makecell[c]{23.7 \\ \textcolor{BetterGreen}{\scriptsize (+14.6)}} 
& \makecell[c]{36.2 \\ \textcolor{BetterGreen}{\scriptsize (+29.1)}} 
& \makecell[c]{15.7 \\ \textcolor{BetterGreen}{\scriptsize (+2.9)}} 
& \makecell[c]{44.3 \\ \textcolor{BetterGreen}{\scriptsize (+35.1)}} 

& \makecell[c]{46.3 \\ \textcolor{BetterGreen}{\scriptsize (-342.4)}} 
& \makecell[c]{34.4 \\ \textcolor{BetterGreen}{\scriptsize (-31.7)}} 
& \makecell[c]{39.9 \\ \textcolor{gray}{\scriptsize (--)}}
& \makecell[c]{20.1 \\ \textcolor{red}{\scriptsize (+14.7)}} 
& \makecell[c]{10.2 \\ \textcolor{BetterGreen}{\scriptsize (-166.4)}} 
\\
\midrule

\multirow{1}{*}{Trapfield} 
& Fine-tuned 
& \makecell[c]{0.0 \\ \textcolor{gray}{\scriptsize (+0.0)}} 
& \makecell[c]{0.0 \\ \textcolor{gray}{\scriptsize (+0.0)}} 
& \makecell[c]{38.9 \\ \textcolor{BetterGreen}{\scriptsize (+38.9)}} 
& \makecell[c]{0.0 \\ \textcolor{gray}{\scriptsize (+0.0)}} 
& \makecell[c]{0.0 \\ \textcolor{gray}{\scriptsize (+0.0)}} 

& \makecell[c]{93.1 \\ \textcolor{BetterGreen}{\scriptsize (+86.2)}} 
& \makecell[c]{40.3 \\ \textcolor{BetterGreen}{\scriptsize (+27.8)}} 
& \makecell[c]{100.0 \\ \textcolor{BetterGreen}{\scriptsize (+100.0)}} 
& \makecell[c]{79.2 \\ \textcolor{BetterGreen}{\scriptsize (+47.3)}} 
& \makecell[c]{6.9 \\ \textcolor{red}{\scriptsize (-4.2)}} 

& \makecell[c]{10.9 \\ \textcolor{BetterGreen}{\scriptsize (+4.3)}} 
& \makecell[c]{12.9 \\ \textcolor{BetterGreen}{\scriptsize (+3.8)}} 
& \makecell[c]{79.1 \\ \textcolor{BetterGreen}{\scriptsize (+72.0)}} 
& \makecell[c]{14.7 \\ \textcolor{BetterGreen}{\scriptsize (+1.9)}} 
& \makecell[c]{10.0 \\ \textcolor{BetterGreen}{\scriptsize (+0.8)}} 

& \makecell[c]{57.5 \\ \textcolor{BetterGreen}{\scriptsize (-331.2)}} 
& \makecell[c]{16.8 \\ \textcolor{BetterGreen}{\scriptsize (-49.3)}} 
& \makecell[c]{3.9 \\ \textcolor{gray}{\scriptsize (--)}}
& \makecell[c]{11.4 \\ \textcolor{red}{\scriptsize (+6.0)}} 
& \makecell[c]{57.8 \\ \textcolor{BetterGreen}{\scriptsize (-118.8)}} 
\\

\bottomrule
\end{tabular}
}
\caption{
Comparison between the baseline (Wan2.2-TI2V-5B) and task-specific fine-tuned models across five game types (Base, Irreg, Trap, 3D, Soko). 
Each cell reports absolute performance and relative change over the baseline on four metrics: EM, SR, PR, and SD.
\textcolor{BetterGreen}{Green} indicates improvement, 
\textcolor{red}{red} indicates degradation, 
and \textcolor{gray}{gray} denotes no change or undefined difference.
}
\label{tab:shared_baseline_performance_makecell}
\end{table*}

In this section, we discuss the observations and insights we
draw from our comprehensive evaluation experiments.

\noindent\textbf{· Wan-R1 Outperforms Prior Models on VR-Bench.} 
As shown in Table~\ref{main_comparison}, our method consistently achieves top performance across nearly all tasks and evaluation metrics, demonstrating both high accuracy and rollout efficiency.
Notably, \textit{Wan-R1} attains a perfect \textit{SR} of \textcolor{BetterGreen}{100.0} on the Trap and 3D maze tasks, highlighting its robust success capabilities even in complex environments.
Compared to its base model \textit{Wan2.2-TI2V-5B$^\Diamond$}, \textit{Wan-R1} achieves a remarkable \textit{EM} improvement of \textcolor{BetterGreen}{+65.3} on 3D, and reduces \textit{SD} by \textcolor{BetterGreen}{100.1} on Soko.
These gains underscore the effectiveness of our fine-tuning strategy in enhancing both correctness and trajectory quality across diverse reasoning settings.

\noindent\textbf{· Success Alone Does Not Guarantee Efficient Reasoning.}
Some models manage to complete tasks, but their rollouts remain highly inefficient.
For instance, \textit{Sora-2} and \textit{MiniMax-Hailuo-2.3} achieve strong \textit{SR} of \textcolor{BetterGreen}{75.0} and \textcolor{BetterGreen}{68.1} on the Base task, yet their corresponding \textit{SD} values reach \textcolor{red}{302.9} and \textcolor{red}{464.0}, revealing substantial path redundancy. Even more striking, the open-source VLM \textit{Qwen2.5-VL-7B} produces a very low \textit{SR} of \textcolor{red}{1.4}, but still yields a high \textit{SD} of \textcolor{red}{300.0}, indicating unstable or erratic generation.
As shown in Table~\ref{main_comparison}, \textit{Wan-R1} achieves a comparable or better \textit{SR} while reducing \textit{SD} to just \textcolor{BetterGreen}{10.3}, demonstrating its ability to generate correct and efficient trajectories consistently.


\noindent\textbf{· Reasoning via Video Outperforms Reasoning via Text.}
Under the same training data and settings, we fine-tune both the vision-language model (\textit{Qwen2.5-VL-7B}) and the video model (\textit{Wan2.2-TI2V-5B}). As shown in Table~\ref{main_comparison}, the video-based model (\textit{Wan-R1}) yields significantly larger gains across all metrics and tasks, especially in challenging settings like Trap and 3D. In contrast, \textit{Qwen2.5-VL-7B-SFT} shows only moderate improvements. This highlights the advantage of \textit{reasoning via video} in learning temporal reasoning and efficient path planning over static VLMs.

\noindent\textbf{· Rule Compliance and Structural Fidelity.} 
As shown in Table~\ref{comparison}, our model \textit{Wan-R1} consistently achieves the highest \textit{VLM-Score} across all maze types, and performs competitively in \textit{MF}, ranking top-2 in most categories: \textbf{\textit{1)}} it consistently attains the highest \textit{VLM-Score} across all maze types (\textcolor{BetterGreen}{$\geq$ 4.0}), indicating superior rule-following behavior in motion continuity, physical plausibility, and environmental interactions; 
\textbf{\textit{2)}} it ranks among the top performers in \textit{MF}, especially excelling on structurally complex mazes like Irreg and 3D. 
\textbf{\textit{3)}} it shows consistent improvements over its base model \textit{Wan2.2-TI2V-5B}, confirming the effectiveness of our training paradigm in enhancing both visual stability and behavioral correctness.

\begin{figure}[h!]
\centering
\includegraphics[width=1\columnwidth]{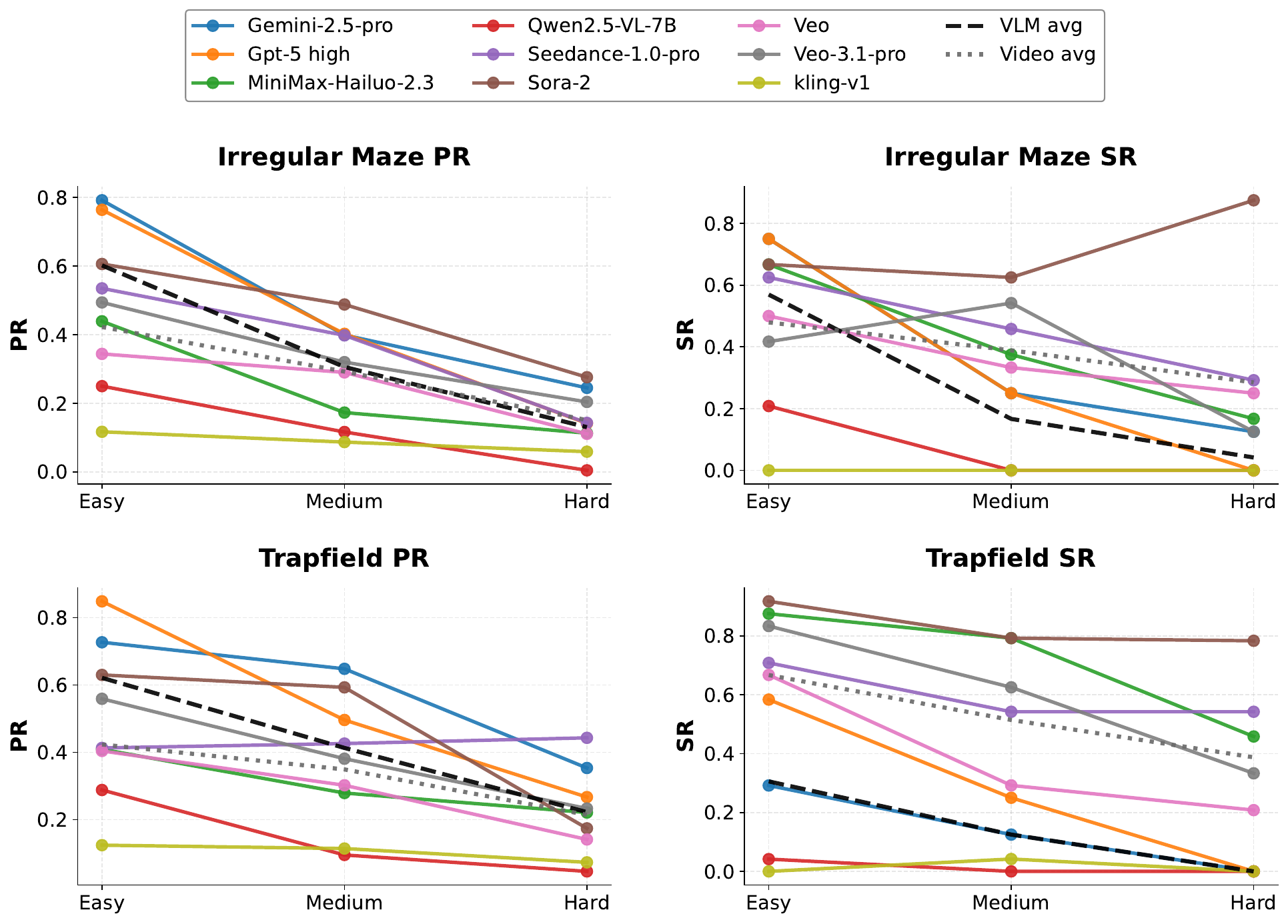} 
\caption{Model performance (PR and SR) on Irregular Maze and Trapfield across difficulty levels. Each curve represents a baseline, while the dashed and dotted lines indicate VLM and Video Model averages. Results for other maze types are in the Appendix.}
\label{fig:benchcompare}
\end{figure}

\noindent\textbf{· Reasoning via Video Scales Better.} 
\textbf{\textit{1)}} As illustrated in Figure~\ref{fig:benchcompare}, model performance consistently declines with increasing maze difficulty on the Trap and Irre Maze tasks. In Easy settings, VLMs often match or even surpass state-of-the-art video models. However, as the maze complexity escalates, VLMs experience a sharper performance drop compared to video models. On large-scale hard mazes, even top-tier VLMs such as \textit{Gemini-2.5-Pro} and \textit{GPT-5} are outperformed by leading video models like \textit{Sora-2} and \textit{Seedance-1.0-pro}. \textbf{\textit{2)}} We attribute this trend to fundamental differences in the reasoning paradigms of the two model families. VLMs rely on encoding static visual observations into textual tokens and performing reasoning within a language-dominant latent space. As maze size increases, the number of visual tokens grows substantially, leading to context-length saturation and degraded long-horizon reasoning. In contrast, video models reason \textit{via visual dynamics}, constructing a temporally grounded CoF that maintains spatial continuity across time. This video-centric reasoning mechanism preserves efficiency as scene complexity increases, since the number of visual tokens per frame remains stable, with visual tokens carrying significantly higher information density than textual tokens, a finding validated by DeepSeek OCR \citep{wei2025deepseekocr} through optical context compression. Remarkably, models such as \textit{Sora-2} even exhibit improved performance in the Irreg Maze under higher difficulty levels, particularly in the \textit{SR} metric.
\textbf{\textit{3)}} These observations suggest that \textit{reasoning via video} constitutes a more native and scalable paradigm for visual reasoning, enabling temporal–spatial problem solving that remains robust under increasing environmental complexity.

\begin{figure}[h!]
\centering
\includegraphics[width=1\columnwidth]{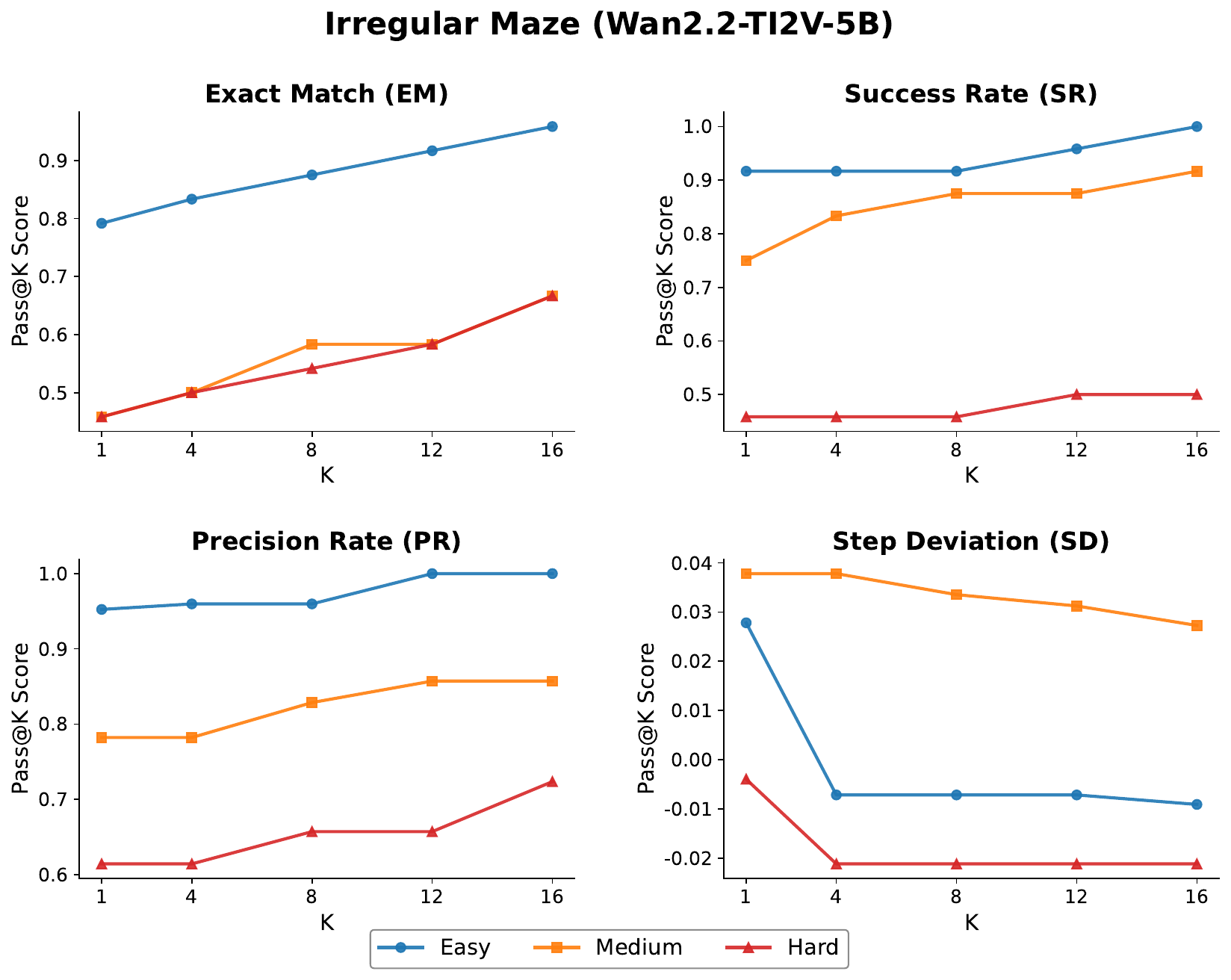} 
\caption{Performance on Irregular Maze using Wan-R1 under test-time scaling. Results are shown across different sampling numbers ($K \in {1,4,8,12,16}$) and difficulty levels (Easy, Medium, Hard). Results for other maze types are in the Appendix.}
\label{fig:irregular-scaling}
\end{figure}

\noindent\textbf{· Test-Time Scaling for Video Models.} 
\textbf{\textit{1)}} Test-time scaling (TTS), exemplified by self-consistency \citep{wang2023selfconsistency}, has shown strong effectiveness in text-based reasoning tasks. Its key intuition is that complex reasoning problems often admit multiple valid solution trajectories, and sampling diverse paths increases the likelihood of converging to a correct answer. Maze-solving naturally shares this property: the solution space is open-ended, and multiple routes can lead to the goal. Motivated by this, we apply TTS to video models by perturbing the sampling noise to generate diverse rollouts, and evaluate performance using Pass@K, which selects the best solution among $K$ independent attempts.
\textbf{\textit{2)}} Figure~\ref{fig:irregular-scaling} shows the scaling behavior of \textit{Wan-R1} on the Irregular Maze benchmark. As $K$ increases from 1 to 16, the model achieves steady gains across all difficulty levels, with improvements of roughly \textcolor{BetterGreen}{10--20\%} depending on metric and difficulty. On Easy mazes, performance rises almost monotonically and nears saturation at higher $K$. Medium difficulty shows clear early gains, often improving by \textcolor{BetterGreen}{5--10\%} from $K=1$ to $K=4$, followed by continued smaller increases. Even on Hard mazes, where Pass@1 scores are lower, TTS still yields consistent upward trends, recovering solutions that single attempts often miss.
\textbf{\textit{3)}} Overall, these results demonstrate that TTS significantly enhances video-model reasoning in VTR tasks. By initializing generation from different noise conditions, the model explores multiple solution pathways within the maze’s open-ended search space. This multi-path exploration effectively unlocks additional reasoning capacity, enabling video models to produce trajectories that are more accurate, more reliable, and consistently closer to the desired solution than those obtained under standard single-sample inference.

\noindent\textbf{· Difficulty Generalization} 
The results in Table \ref{tab:difficulty_single} demonstrate the strong difficulty generalization capability of \textit{Wan-R1} across all five maze tasks. Although the model is fine-tuned only on the Easy level, it consistently delivers substantial improvements over the \textit{Wan2.2-TI2V-5B} baseline on Medium and Hard mazes as well. This indicates that \textit{Wan-R1} does not simply memorize small or low-complexity layouts; rather, it internalizes a more principled and transferable reasoning procedure. The gains observed across unseen difficulty tiers show that fine-tuning on small mazes induces broad generalization: the model acquires a maze-solving strategy that scales to larger, more intricate structures without additional supervision. Such behavior reflects a deeper structural understanding of the environment and verifies that \textit{Wan-R1}’s improvements stem from reasoning-pattern internalization rather than task-specific overfitting.

\noindent\textbf{· Maze Type Generalization.} 
As shown in Table~\ref{tab:shared_baseline_performance_makecell}, fine-tuning on a single game (e.g., \textit{Regular Maze}, \textit{Trapfield}, or \textit{3D Maze}) not only improves in-domain performance but also yields substantial gains on unseen games across all metrics (\textit{EM}, \textit{SR}, \textit{PR}, \textit{SD}). This highlights the emergence of transferable video reasoning capabilities. Notably, models fine-tuned on \textit{3D Maze} exhibit strong generalization, this overall transfer pattern underscores that training on complex 3D structures fosters general reasoning skills applicable to other maze types.

\noindent\textbf{· Texture Generalization.}
Although fine-tuned only on the \textit{Raw} skin of each game type, the model shows consistent and often substantial gains on unseen textures (\textit{Skin2} and \textit{Skin3}). As shown in Table~\ref{tab:texture}, this pattern holds across all five task domains. For example, in the Base task, Skin3—never encountered during training—improves by \textcolor{BetterGreen}{+23.6} in \textit{EM} and \textcolor{BetterGreen}{+41.6} in \textit{PR}. In the more challenging 3D Maze, the generalization is even stronger, reaching \textcolor{BetterGreen}{+50.0} \textit{EM} and \textcolor{BetterGreen}{+78.5} \textit{PR} on Skin3. These results demonstrate the model’s strong texture-level generalization and indicate that the learned spatio-temporal reasoning transfers robustly to new visual styles.

\vspace{-8pt}

\begin{table}[t]
\centering
\resizebox{\linewidth}{!}{
\begin{tabular}{l|ccc|ccc|ccc|ccc}
\toprule
\textbf{Task} 
& \multicolumn{3}{c|}{\textbf{EM}} 
& \multicolumn{3}{c|}{\textbf{SR}} 
& \multicolumn{3}{c|}{\textbf{PR}} 
& \multicolumn{3}{c}{\textbf{SD}} \\
& Raw & Skin2 & Skin3 & Raw & Skin2 & Skin3 & Raw & Skin2 & Skin3 & Raw & Skin2 & Skin3 \\
\midrule

\multirow{3}{*}{Base}
& 0.0 & 0.0 & 0.0 
& 6.9 & 4.2 & 2.8
& 6.6 & 4.6 & 9.4
& 388.7 & 11.7 & 14.9 \\

& 33.3 & 1.4 & 23.6 
& 76.4 & 38.9 & 68.1 
& 60.6 & 12.3 & 51.0 
& 10.3 & 19.0 & 9.0 \\

& \makecell[c]{\textcolor{BetterGreen}{\scriptsize (+33.3)}} 
& \makecell[c]{\textcolor{BetterGreen}{\scriptsize (+1.4)}} 
& \makecell[c]{\textcolor{BetterGreen}{\scriptsize (+23.6)}} 
& \makecell[c]{\textcolor{BetterGreen}{\scriptsize (+69.5)}} 
& \makecell[c]{\textcolor{BetterGreen}{\scriptsize (+34.7)}} 
& \makecell[c]{\textcolor{BetterGreen}{\scriptsize (+65.3)}} 
& \makecell[c]{\textcolor{BetterGreen}{\scriptsize (+54.0)}} 
& \makecell[c]{\textcolor{BetterGreen}{\scriptsize (+7.7)}} 
& \makecell[c]{\textcolor{BetterGreen}{\scriptsize (+41.6)}} 
& \makecell[c]{\textcolor{BetterGreen}{\scriptsize (-378.4)}} 
& \makecell[c]{\textcolor{red}{\scriptsize (+7.3)}} 
& \makecell[c]{\textcolor{BetterGreen}{\scriptsize (-5.9)}} 
\\
\midrule

\multirow{3}{*}{Irreg}
& 0.0 & 0.0 & 0.0
& 12.5 & 4.2 & 2.1
& 9.1 & 12.0 & 47.8
& 66.1 & 39.5 & 42.0 \\

& 56.9 & 22.2 & 15.3 
& 69.4 & 15.3 & 23.6 
& 71.6 & 36.5 & 26.2 
& 2.4 & 5.1 & 7.7 \\

& \makecell[c]{\textcolor{BetterGreen}{\scriptsize (+56.9)}}
& \makecell[c]{\textcolor{BetterGreen}{\scriptsize (+22.2)}}
& \makecell[c]{\textcolor{BetterGreen}{\scriptsize (+15.3)}}
& \makecell[c]{\textcolor{BetterGreen}{\scriptsize (+56.9)}}
& \makecell[c]{\textcolor{BetterGreen}{\scriptsize (+11.1)}}
& \makecell[c]{\textcolor{BetterGreen}{\scriptsize (+21.5)}}
& \makecell[c]{\textcolor{BetterGreen}{\scriptsize (+62.5)}}
& \makecell[c]{\textcolor{BetterGreen}{\scriptsize (+24.5)}}
& \makecell[c]{\textcolor{red}{\scriptsize (-21.6)}}
& \makecell[c]{\textcolor{BetterGreen}{\scriptsize (-63.7)}}
& \makecell[c]{\textcolor{BetterGreen}{\scriptsize (-34.4)}}
& \makecell[c]{\textcolor{BetterGreen}{\scriptsize (-34.3)}}
\\
\midrule

\multirow{3}{*}{3D}
& 0.0 & 0.0 & 0.0
& 0.0 & 4.2 & 4.2
& 7.1 & 8.9 & 8.3
& -- & 73.6 & 59.3 \\

& 65.3 & 43.1 & 50 
& 100.0 & 97.2 & 100 
& 93.5 & 83.8 & 86.8  
& 3.9 & 6.4 & 5.9   \\

& \makecell[c]{\textcolor{BetterGreen}{\scriptsize (+65.3)}} 
& \makecell[c]{\textcolor{BetterGreen}{\scriptsize (+43.1)}} 
& \makecell[c]{\textcolor{BetterGreen}{\scriptsize (+50.0)}}
& \makecell[c]{\textcolor{BetterGreen}{\scriptsize (+100.0)}} 
& \makecell[c]{\textcolor{BetterGreen}{\scriptsize (+93.0)}} 
& \makecell[c]{\textcolor{BetterGreen}{\scriptsize (+95.8)}}
& \makecell[c]{\textcolor{BetterGreen}{\scriptsize (+86.4)}} 
& \makecell[c]{\textcolor{BetterGreen}{\scriptsize (+74.9)}} 
& \makecell[c]{\textcolor{BetterGreen}{\scriptsize (+78.5)}}
& \makecell[c]{\textcolor{gray}{\scriptsize (--)}}
& \makecell[c]{\textcolor{BetterGreen}{\scriptsize (-67.2)}} 
& \makecell[c]{\textcolor{BetterGreen}{\scriptsize (-53.4)}} 
\\
\midrule

\multirow{3}{*}{Soko}
& 0.0 & 0.0 & 0.0
& 31.9 & 2.8 & 7.1
& 12.8 & 9.3 & 8.1
& 5.4 & -- & -- \\

& 4.2 & 0.0 & 1.4 
& 69.4 & 34.7 & 58.3 
& 44.3 & 21.0 & 14.1 
& 10.2 & 58.6 & 82.6 \\

& \makecell[c]{\textcolor{BetterGreen}{\scriptsize (+4.2)}} 
& \makecell[c]{\textcolor{gray}{\scriptsize (+0.0)}} 
& \makecell[c]{\textcolor{BetterGreen}{\scriptsize (+1.4)}}
& \makecell[c]{\textcolor{BetterGreen}{\scriptsize (+37.5)}} 
& \makecell[c]{\textcolor{BetterGreen}{\scriptsize (+31.9)}} 
& \makecell[c]{\textcolor{BetterGreen}{\scriptsize (+51.2)}}
& \makecell[c]{\textcolor{BetterGreen}{\scriptsize (+31.5)}} 
& \makecell[c]{\textcolor{BetterGreen}{\scriptsize (+11.7)}} 
& \makecell[c]{\textcolor{BetterGreen}{\scriptsize (+6.0)}}
& \makecell[c]{\textcolor{red}{\scriptsize (+4.8)}} 
& \makecell[c]{\textcolor{gray}{\scriptsize (--)}}
& \makecell[c]{\textcolor{gray}{\scriptsize (--)}}
\\
\midrule

\multirow{3}{*}{Trap}
& 0.0 & 0.0 & 0.0
& 11.1 & 0.0 & 0.0
& 9.2 & 9.0 & 7.5
& 176.6 & -- & -- \\

& 38.9 & 9.7 & 0.0 
& 100.0 & 38.9 & 1.4 
& 79.1 & 29.0 & 9.9 
& 3.9 & 8.7 & 18.5 \\

& \makecell[c]{\textcolor{BetterGreen}{\scriptsize (+38.9)}} 
& \makecell[c]{\textcolor{BetterGreen}{\scriptsize (+9.7)}} 
& \makecell[c]{\textcolor{gray}{\scriptsize (+0.0)}}
& \makecell[c]{\textcolor{BetterGreen}{\scriptsize (+88.9)}} 
& \makecell[c]{\textcolor{BetterGreen}{\scriptsize (+38.9)}} 
& \makecell[c]{\textcolor{BetterGreen}{\scriptsize (+1.4)}}
& \makecell[c]{\textcolor{BetterGreen}{\scriptsize (+69.9)}} 
& \makecell[c]{\textcolor{BetterGreen}{\scriptsize (+20.0)}} 
& \makecell[c]{\textcolor{BetterGreen}{\scriptsize (+2.4)}}
& \makecell[c]{\textcolor{BetterGreen}{\scriptsize (-172.7)}} 
& \makecell[c]{\textcolor{gray}{\scriptsize (--)}}
& \makecell[c]{\textcolor{gray}{\scriptsize (--)}}
\\

\bottomrule
\end{tabular}}
\caption{
Texture generalization under different skin. 
For each task, the table reports the baseline performance, the results after fine-tuning on the Raw texture, 
and the relative change across three texture conditions (Raw, Skin2, Skin3) on EM, SR, PR, and SD. 
\textcolor{BetterGreen}{Green} denotes improvement, 
\textcolor{red}{red} denotes degradation, 
and \textcolor{gray}{gray} indicates no change or undefined differences.
}
\label{tab:texture}
\end{table}


\section{Conclusion}

In this work, we take a step forward in evaluating whether video models can reason via video generation. We propose \textit{\textbf{VR-Bench}}, a comprehensive benchmark grounded in maze-solving tasks to assess the spatial reasoning ability of video models. Our experiments demonstrate that fine-tuned video models exhibit strong spatial reasoning and consistently outperform leading vision-language models. Moreover, our analysis reveals a test-time scaling effect akin to self-consistency in language models, underscoring the scalable potential of video-based reasoning.

\noindent\textbf{Limitations and Future Work.} While VR-Bench provides a focused and rigorous testbed for spatial reasoning, it currently emphasizes maze-centric tasks. Future iterations of VR-Bench will explore broader and more challenging reasoning scenarios. For instance, we plan to incorporate Olympiad-level problem-solving tasks, such as solving complex physics or mathematics competition problems via video-based visual reasoning. In addition, we aim to support embodied reasoning settings where models are required to predict or simulate coherent action sequences within interactive environments.

{
    \small
    \bibliographystyle{ieeenat_fullname}
    \bibliography{main}
}

\clearpage
\setcounter{page}{1}
\maketitlesupplementary

\section{Experiment Details}

\subsection{Implementation Details of Baselines} 

\paragraph{Video Models.}
To ensure fair comparison across different video models, we standardized input preprocessing and output postprocessing. Since most maze images are square ($1:1$), we applied black-border padding to satisfy model-specific resolution or aspect ratio requirements, followed by center cropping to restore the maze region. Generated videos were temporally aligned according to each model’s fixed or adjustable duration settings.

Veo-3.1 and Veo-3.1-Pro generate fixed 8-second videos in a $9:16$ aspect ratio, with black-border padding and post-cropping applied.
Doubao outputs 10-second, $1:1$ videos, padding and cropping inputs smaller than $300\times300$ pixels.
Kling produces 10-second videos without further adjustments.
MiniMax also yields 10-second videos at its default 768p resolution, using the same padding–cropping scheme for small inputs.
Sora-2 generates 10-second, $9:16$ videos, followed by black-border padding and cropping to maintain maze integrity.

\paragraph{VLM.}  
Given an initial observation image $I_0$, the VLM predicts an action sequence $a_{\text{pred}} = [a_1, \dots, a_T]$, representing its intended movements in the environment. The actions are sequentially executed in the simulator to verify trajectory validity against the optimal reference $a_{\text{opt}}$.

For each task type, we define its corresponding action space. In TrapField, Sokoban, and Regular Maze, actions correspond to four-directional moves \{up, down, left, right\}, and the VLM outputs sequences such as [``up'', ``right'', ``down'']. In the Irregular Maze, actions are defined over irregular graph nodes (A, B, \dots), where the VLM outputs node transition sequences like [``A'', ``C'', ``E'']; the model is evaluated on whether these transitions correspond to valid path connections. For the 3D Maze, actions include six directional movements covering both horizontal and vertical axes, and the VLM outputs sequences such as [``forward\_left'', ``up'', ``forward\_right''].

\subsection{Training Details} 

We fine-tuned one model for each of the five maze categories in our benchmark.
For each category, the fine-tuning set consists of 80\% of the data from the
first skin, covering the \textit{easy}, \textit{medium}, and \textit{hard} difficulty levels.
All models were trained using the DiffSynth-Studio framework.
The training hyperparameters are summarized in Table~\ref{tab:training_params}.

\begin{table}[t]
\centering
\small
\begin{tabular}{ll}
\toprule
\textbf{Parameter} & \textbf{Value} \\
\midrule
Frame resolution    & $512 \times 512$ \\
Number of frames    & 193 \\
Dataset repeat factor & 100 \\
Base model          & Wan2.2-TI2V-5B \\
LoRA rank           & 32 \\
Learning rate       & $1 \times 10^{-4}$ \\
Epochs              & 5 \\
\bottomrule
\end{tabular}
\caption{Key training parameters used in our fine-tuning setup.}
\label{tab:training_params}
\end{table}

\begin{figure}[h!]
\centering
\includegraphics[width=1\columnwidth]{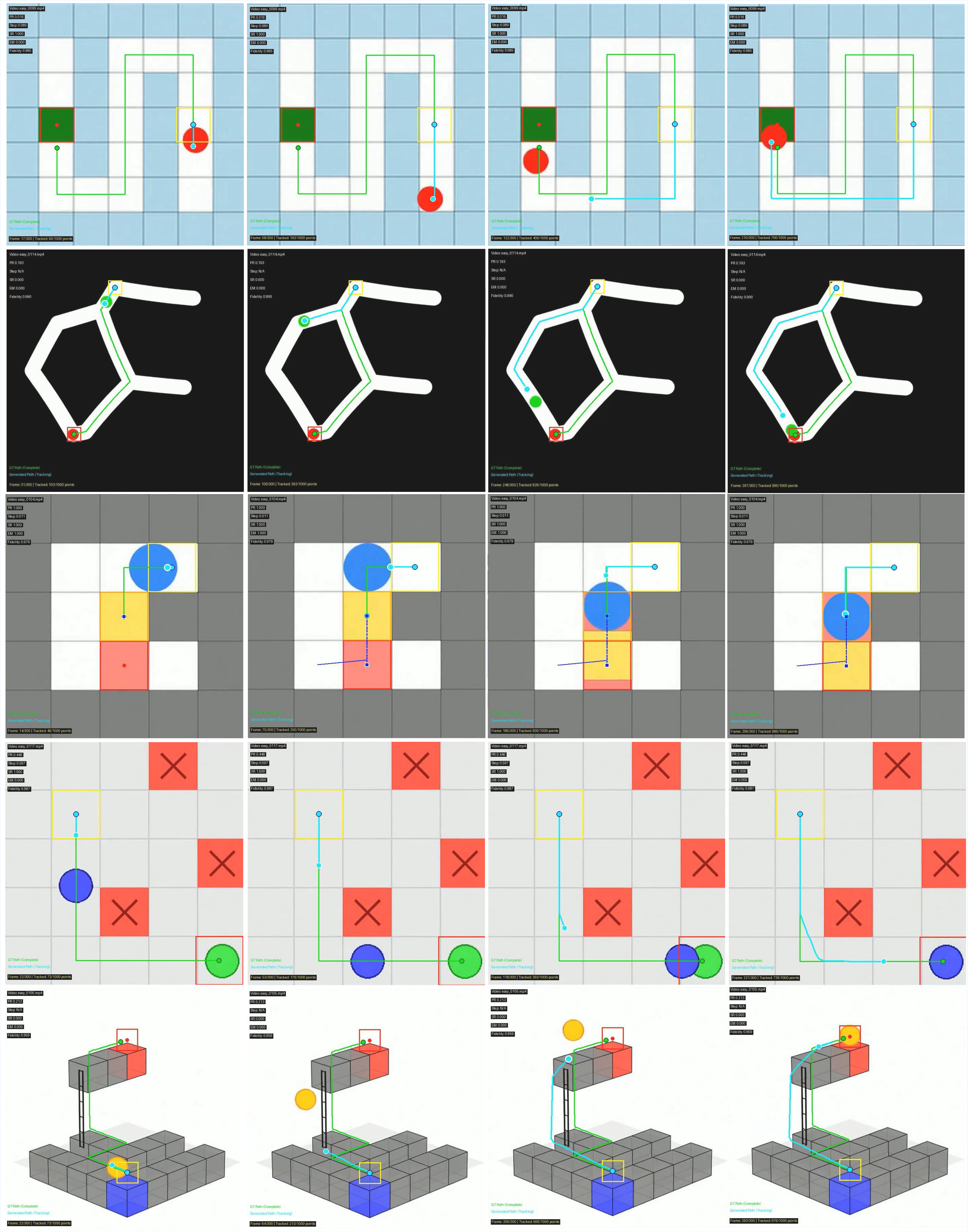} 
\caption{Dynamic visualization of trajectory tracking across five maze types.
For each task, the green polyline denotes the ground-truth trajectory, while the blue polyline represents the trajectory tracked from the generated video.
Each column shows a temporally ordered frame sampled from the tracking process, illustrating how the agent’s motion evolves over time.}
\label{fig:vary}
\end{figure}

\section{Evaluation Details} 
\subsection{Path Matching} 
\paragraph{Setup}
For each generated video clip we load its UnifiedState description and enumerate all candidate ground-truth (GT) videos sharing the same scene identifier. The generated clip’s spatial resolution and frame rate define a unified evaluation specification; both generated and GT videos are resized and temporally resampled to this specification before comparison.

\paragraph{Trajectory Extraction}
The controllable agent’s initial bounding box is read from the state and scaled to the evaluation resolution. An object tracker (priority order: CSRT, with fallbacks KCF/MOSSE) is initialized on the first frame. Tracking proceeds frame by frame; sampling follows a fixed temporal interval derived from the unified frame rate. When tracking fails on a sampling frame, the last valid center is reused to preserve temporal alignment. The output is a pixel-space polyline of agent centers across time, paired with the first frame for later visualization.

\paragraph{Normalization and Resampling}
Extracted pixel trajectories are normalized to the unit square using the evaluated video’s width and height. Physical-distance resampling is applied: the GT trajectory is resampled into a fixed number of equidistant points along arc length, and the generated trajectory is interpolated at the same cumulative distances (clipped to its total length). This produces speed-invariant, sparsity-controlled trajectory pairs.

\paragraph{GT Selection and Outputs}
Each generated clip is evaluated against all candidate GT trajectories using a score based solely on trajectory length consistency; the GT with the highest score is selected as the match. For every generated clip, we record the matched GT identity and per-clip summary statistics, and we produce visual diagnostics that overlay GT (green) and generated (blue) resampled trajectories on the first frame, along with start (yellow) and goal (red) bounding boxes. Batch evaluation yields per-difficulty summaries and overall totals, enabling direct comparison of trajectory performance across difficulty levels.

\begin{figure*}[t]
\centering
\includegraphics[width=2\columnwidth]{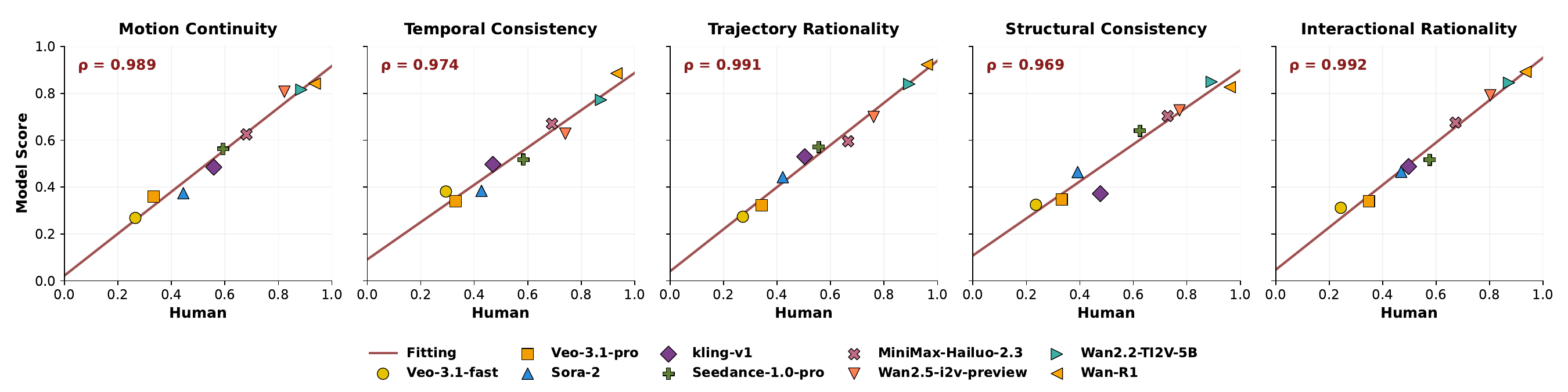}
\caption{
\textbf{Human Alignment of VLM-as-Judge Evaluation.}
Each plot corresponds to Motion Continuity, Temporal Consistency, Trajectory Rationality, Structural Consistency, or Interactional Rationality.
Dots show the human preference win ratio (horizontal axis) and the \textbf{VLM-score} (vertical axis) for nine video generation models.
A linear fit visualizes the correlation, and Pearson’s correlation coefficient ($\rho$) is reported for each dimension, indicating close alignment between VLM-as-Judge scores and human judgment across all five dimensions.
}
\label{fig:align}
\end{figure*}

\subsection{Rule Compliance}


\paragraph{Human Alignment}

To verify that our \textit{VLM-as-Judge} evaluation faithfully reflects human judgment across all five diagnostic dimensions, we conducted a large-scale human preference study over a diverse set of generated videos. The annotation protocol follows standard practices used in prior video-evaluation benchmarks, with controlled instructions, representative examples, and multiple rounds of cross-checking to ensure consistent and reliable human annotations.

In line with established evaluation frameworks, we computed the correlation between the VLM-as-Judge scores and human preference outcomes. Figure~\ref{fig:align} presents the alignment plots, showing the relationship between human win ratios and our VLM-based scores for each dimension, including Motion Continuity, Temporal Consistency, Trajectory Rationality, Structural Consistency, and Interactional Rationality. A linear regression is fitted for visualization, and the Pearson correlation coefficient ($\rho$) is reported for each dimension. Across all five dimensions, VLM-score aligns closely with human judgment, demonstrating that our proposed evaluation protocol provides a reliable and human-consistent assessment of video reasoning quality.

\begin{figure*}[t]
\centering
\includegraphics[width=2\columnwidth]{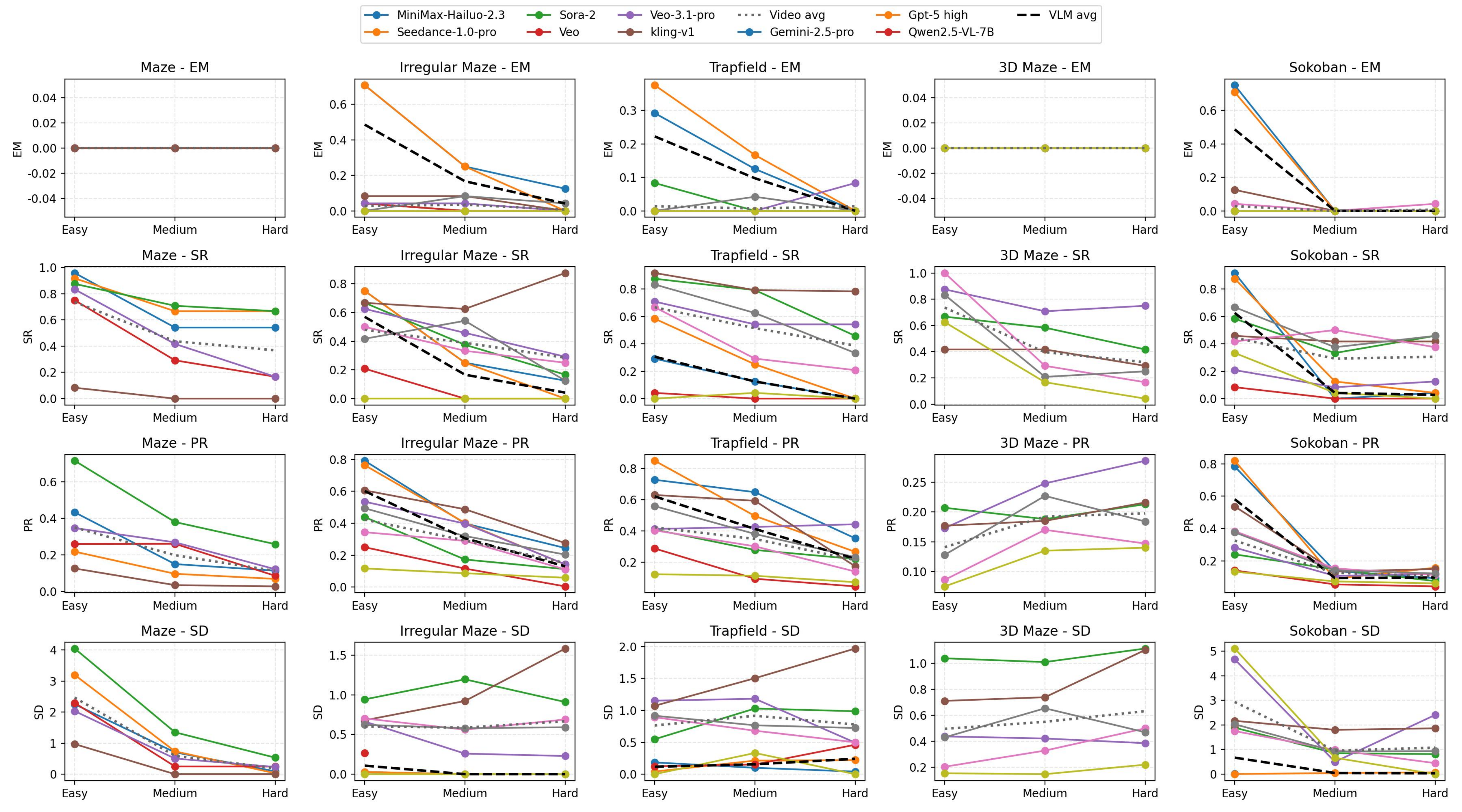}
\caption{
\textbf{Model performance across all five maze types and difficulty levels.}
Each subplot corresponds to a specific maze type (Base, Irregular, Trapfield, 3D, Sokoban) and evaluation metric (Exact Match, Success Rate, Precision Rate, or Step Deviation). Each curve represents a baseline model, while the dashed and dotted lines denote the VLM and Video Model averages. Full metric trends across difficulty levels illustrate the degradation patterns and performance differences among models.}
\label{fig:all_bench}
\end{figure*}

\section{Additional Analysis}

\paragraph{Test-Time Scaling}

\begin{figure*}[t]
\centering

\begin{subfigure}[t]{0.48\textwidth}
    \centering
    \includegraphics[width=\linewidth]{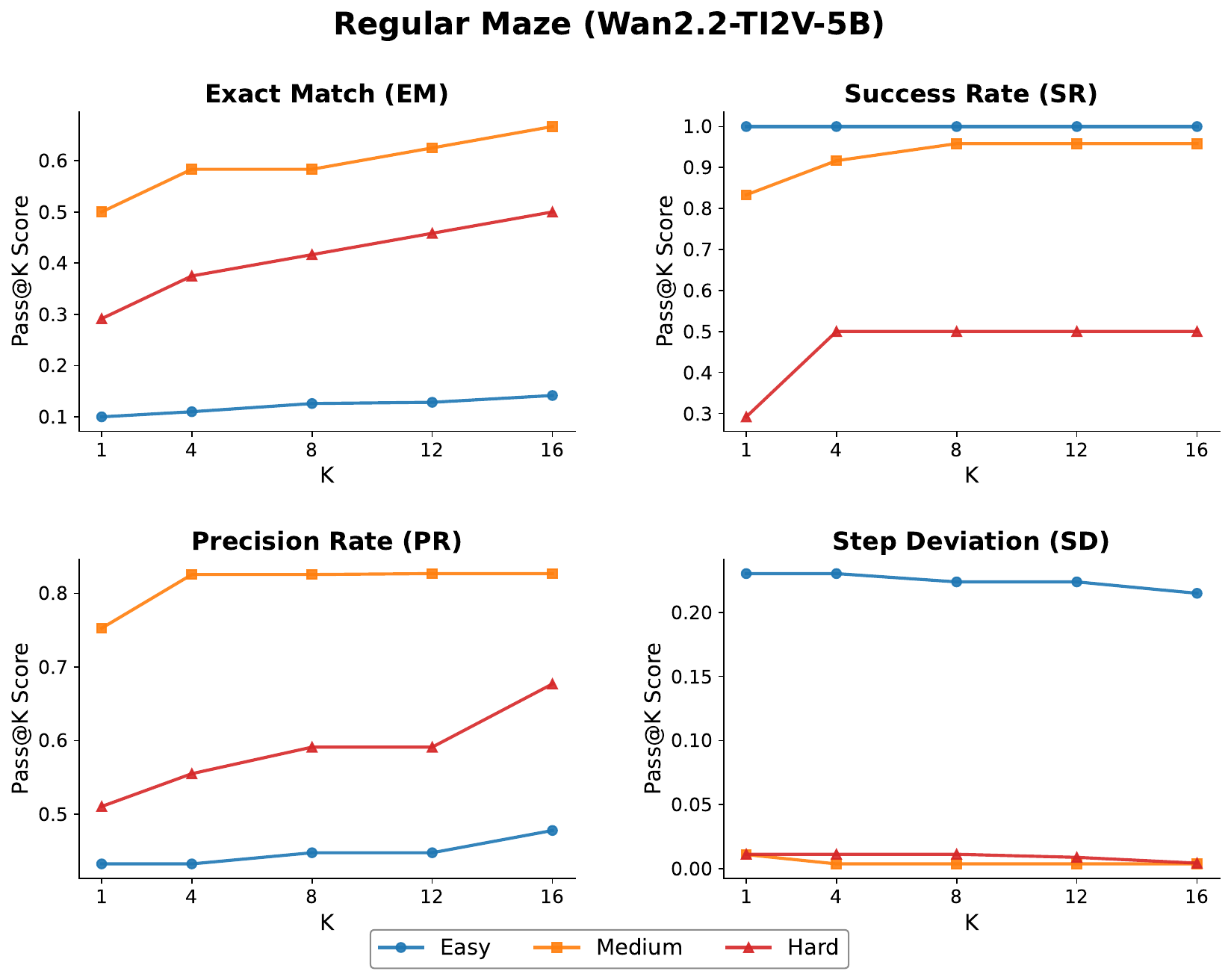}
\end{subfigure}
\hfill
\begin{subfigure}[t]{0.48\textwidth}
    \centering
    \includegraphics[width=\linewidth]{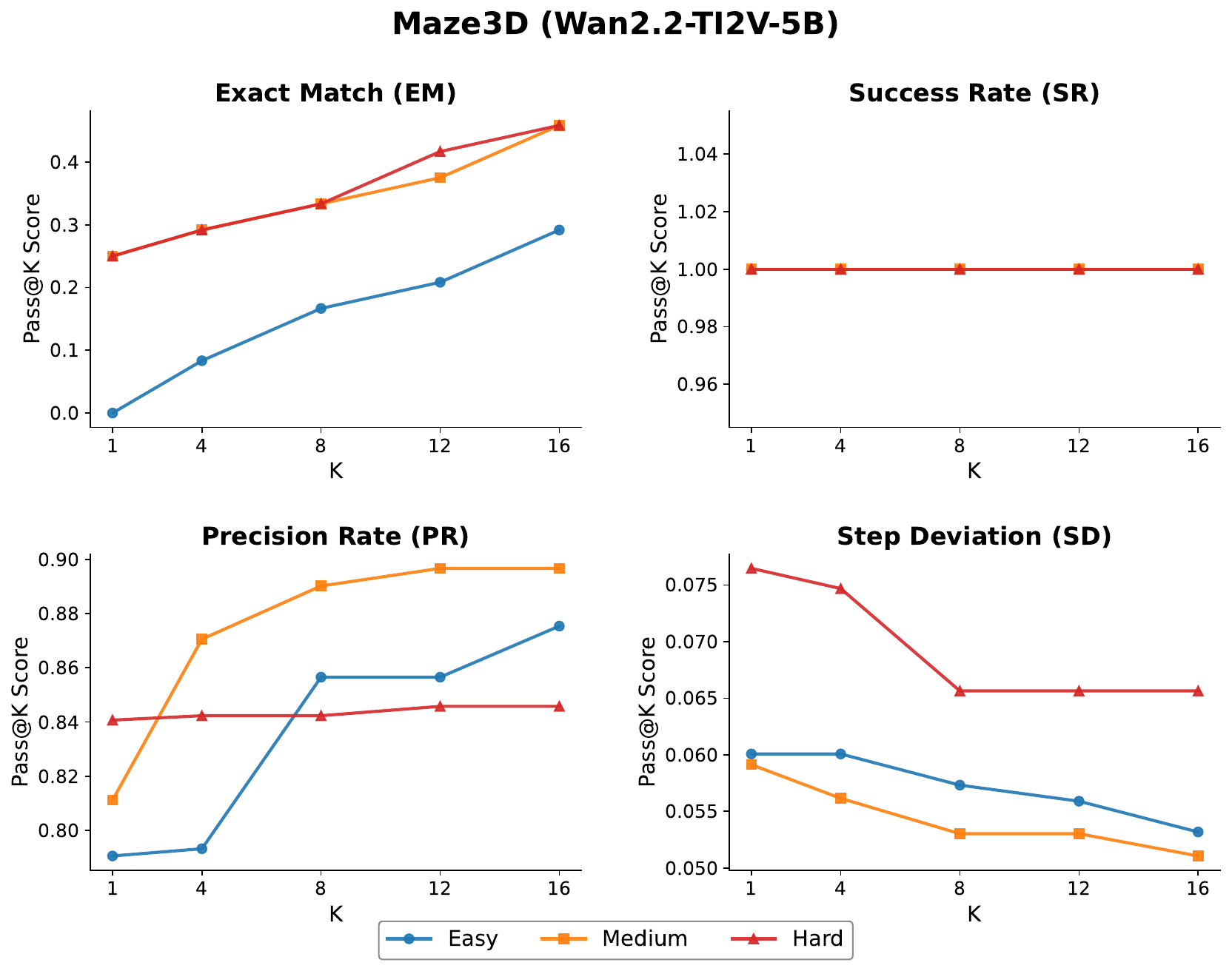}
\end{subfigure}

\vspace{0.5em}

\begin{subfigure}[t]{0.48\textwidth}
    \centering
    \includegraphics[width=\linewidth]{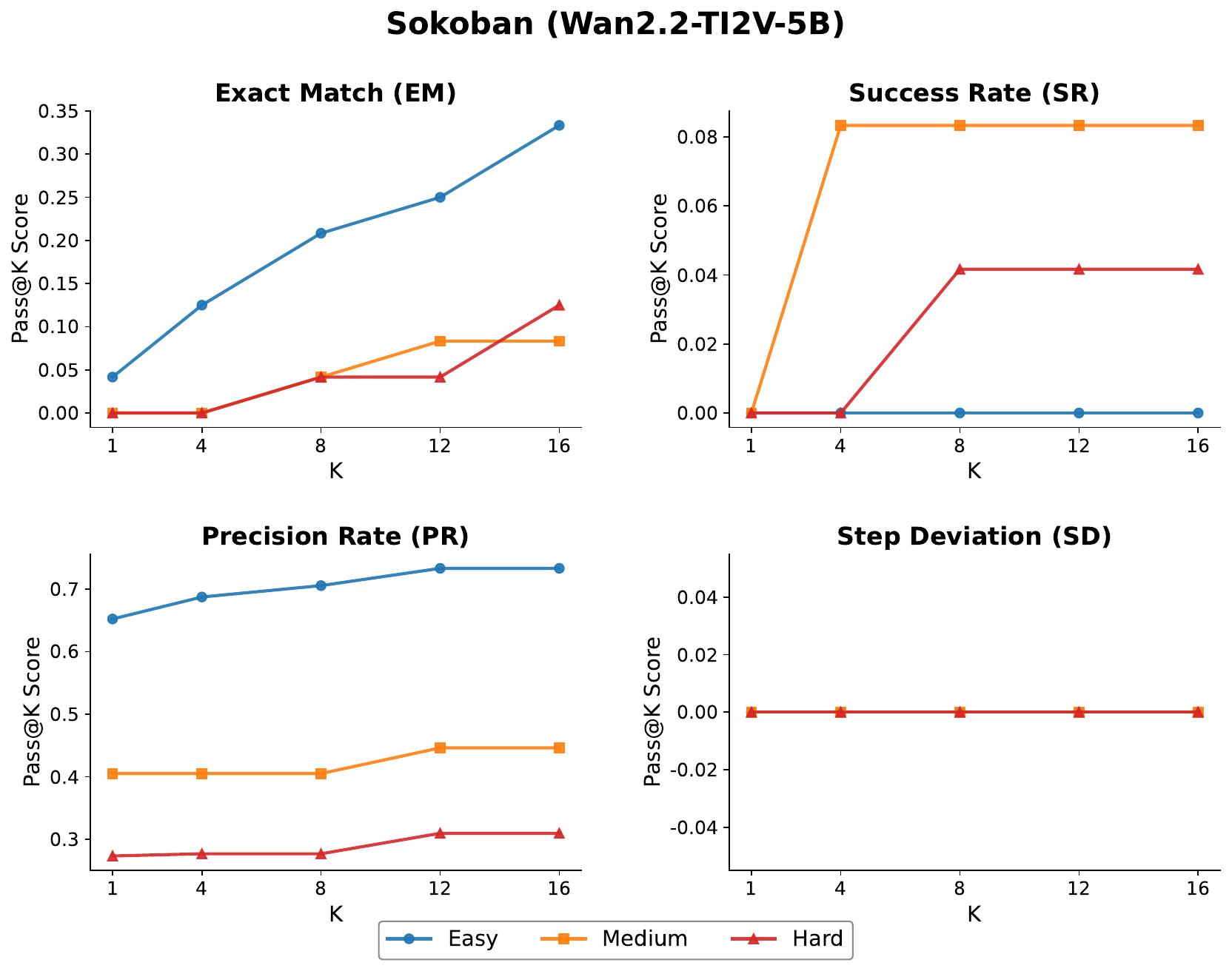}
\end{subfigure}
\hfill
\begin{subfigure}[t]{0.48\textwidth}
    \centering
    \includegraphics[width=\linewidth]{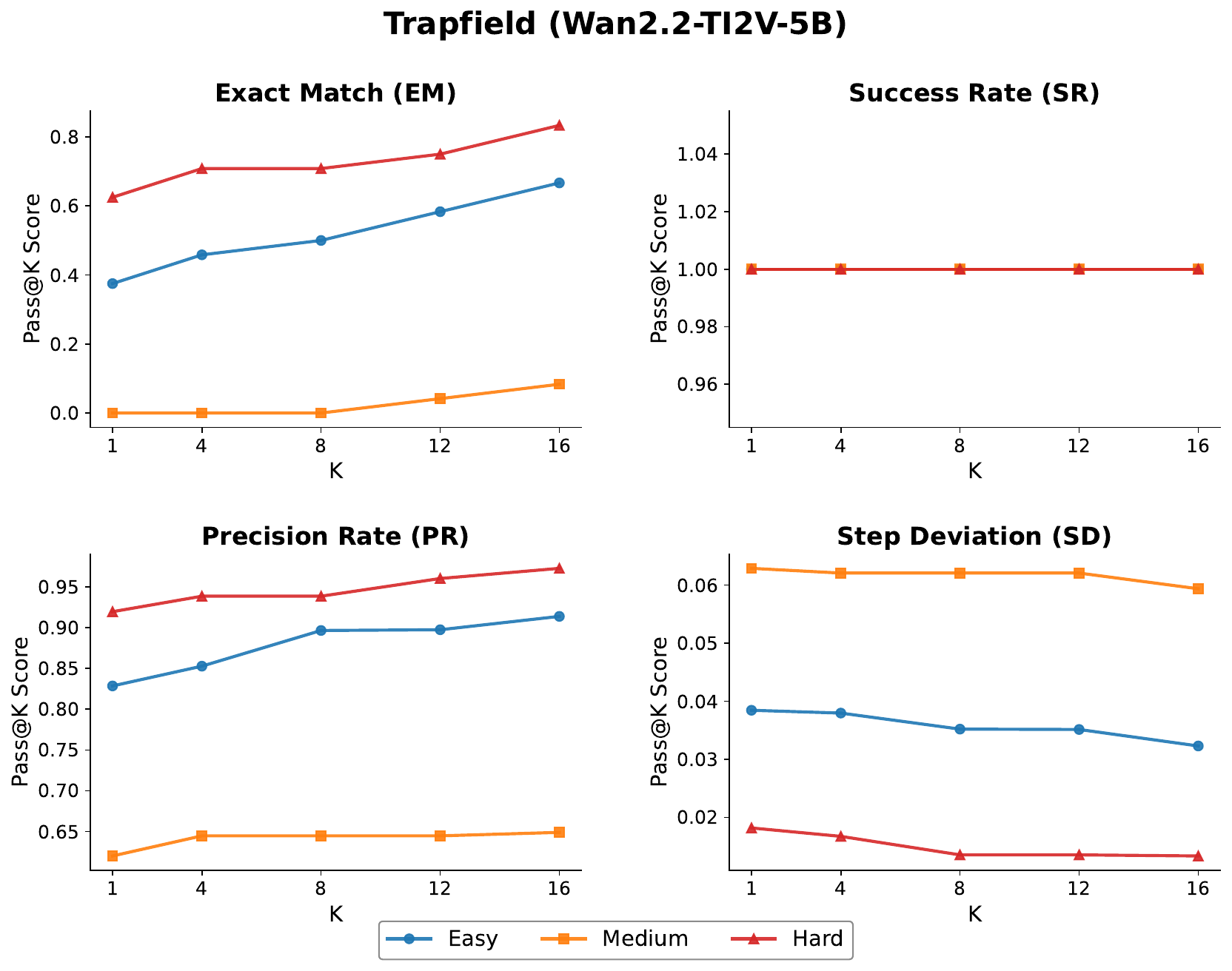}
\end{subfigure}

\caption{
\textbf{Performance across four maze types using Wan-R1 under test-time scaling.}
Results are shown across different sampling numbers ($K \in \{1,4,8,12,16\}$) and difficulty levels (Easy, Medium, Hard). Each subplot corresponds to one maze type (Regular Maze, Irregular Maze, Trapfield, and Sokoban) and one evaluation metric (Exact Match, Success Rate, Precision Rate, or Step Deviation), illustrating how test-time scaling influences trajectory accuracy and stability across diverse maze structures.
}
\label{fig:all_test_time}
\end{figure*}

We evaluate how increasing the number of test-time trajectory samples $K \in \{1, 4, 8, 12, 16\}$ influences model performance across different maze types and difficulty levels. As illustrated in Figure~\ref{fig:all_test_time}, larger $K$ values generally lead to improved performance across four evaluation metrics: Exact Match (EM), Success Rate (SR), Precision Rate (PR), and Step Deviation (SD).

For \textbf{Regular Maze}, performance improvements are consistent and significant across difficulty levels. All metrics—especially EM and PR—demonstrate strong upward trends with increasing $K$, particularly for hard instances, indicating enhanced trajectory fidelity and goal alignment.

In \textbf{Maze3D}, higher $K$ yields substantial gains on EM and PR across all difficulty levels. Notably, SR remains saturated (close to 1.0), suggesting that while reaching the goal is easy, fine-grained precision benefits from more diverse samples. SD decreases with $K$, highlighting trajectory smoothness gains.

\textbf{Sokoban} exhibits the most challenging dynamics: absolute scores remain low, especially on EM and PR. However, performance steadily improves as $K$ increases, most noticeably on the easy setting. This reflects the high complexity introduced by object manipulation and interaction constraints.

For \textbf{Trapfield}, metrics show moderate but consistent gains, with a gradual rise in EM and PR as $K$ increases. SR is already saturated at lower $K$ values, and SD maintains stability, implying that while high-level goals are achievable, detailed path planning benefits from greater sample diversity.

Overall, the results suggest that test-time scaling is an effective strategy for enhancing model robustness, particularly in structurally complex or interaction-heavy environments. The diminishing returns observed on easy levels also motivate future directions such as difficulty-adaptive sampling strategies.

\paragraph{Model Performance Across Difficulty Levels}
To better understand model behavior under varying spatial structures and difficulty levels, we conduct a comprehensive analysis over all five maze types using all four evaluation metrics (EM, SR, PR, SD). As shown in Fig.~\ref{fig:all_bench}, this section highlights global performance trends, examines cross-model and cross-metric differences, and discusses how structural characteristics of each maze family amplify specific failure modes. Together, these observations provide a detailed view of the strengths and limitations of both video models and VLMs in long-horizon trajectory reasoning tasks.

\textbf{1) Overall Difficulty Trends.}
\textit{Performance consistently declines with increasing difficulty across all maze types and metrics.}
Across all five maze types (Base, Irregular, Trapfield, 3D, Sokoban), model performance decreases steadily from Easy to Hard difficulty, and this pattern is consistent across EM, SR, PR, and SD. The decline is most pronounced in Irregular and Trapfield, where complex geometry and branching paths increase long-horizon planning difficulty. On Easy mazes, several VLMs perform comparably to video models, but their performance drops much faster as maze complexity increases. On Hard mazes, even strong VLMs such as \textit{Gemini-2.5-pro} and \textit{GPT-5-high} are frequently surpassed by video models like \textit{Sora-2} and \textit{Seedance-1.0-pro}. In contrast, Base and Sokoban show milder degradation: Base Maze remains structurally simple, allowing more stable PR and SR, while Sokoban’s grid-based layout reduces positional ambiguity despite strict EM constraints.

\textbf{2) Cross-Metric and Cross-Model Patterns.}
\textit{Video models display stronger robustness, while VLMs show high variance across maze structures.}
Across metrics, video models such as \textit{Sora-2}, \textit{Seedance-1.0-pro}, and \textit{Veo-3.1-pro} consistently rank highest, especially in PR and SR, which rely on spatial alignment and correct goal attainment. VLMs including \textit{Gemini-2.5-pro}, \textit{Qwen2.5-VL-7B}, and \textit{GPT-5-high} exhibit large performance variance: they perform well on simpler layouts like Base Maze but degrade sharply on Trapfield and 3D Maze. The difference becomes particularly apparent in SD, where VLM trajectories drift substantially as maze complexity grows. EM remains the strictest metric, and most models fail to achieve meaningful EM on Hard variants of Irregular, Trapfield, 3D Maze, and Sokoban, reflecting the difficulty of producing fully correct long-horizon trajectories.

\textbf{3) Structural Effects Across Maze Families.}
\textit{Different maze structures stress models in distinct ways, highlighting complementary challenges.}
Irregular Maze and Trapfield serve as the strongest discriminators of model capability: only a few video models maintain usable SR and PR at Hard difficulty, while most VLMs collapse almost entirely. 3D Maze introduces challenges related to depth perception and occlusion, resulting in instability in PR and SD even at Medium difficulty. Sokoban emphasizes rule consistency: minor violations lead directly to failure, causing sharp drops in EM and SR on Hard difficulty. 

Overall, these results show that video models maintain stronger spatial-temporal consistency as difficulty increases, whereas VLMs perform well primarily when maze structures are simple or when short-horizon reasoning dominates. The diversity of maze structures provides a comprehensive stress test for evaluating trajectory fidelity.

\section{Failure cases}
To illustrate typical failure cases, we adopt a three-frame visualization format for each example:

\begin{itemize}
    \item \textbf{Left:} The initial frame of the reasoning video, showing the static scene before any agent action.
    \item \textbf{Middle:} A representative intermediate frame capturing the failure behavior during the reasoning process.
    \item \textbf{Right:} A trajectory visualization comparing the predicted and ground-truth reasoning paths. The \textcolor{BetterGreen}{green line} denotes the ground-truth trajectory, the \textcolor{BetterBlue}{blue line} represents the model’s predicted path, the \textcolor{BetterYellow}{yellow square} indicates the starting point, and the \textcolor{BetterRed}{red square} marks the final destination.
\end{itemize}

\begin{figure}[h!]
\centering
\includegraphics[width=1\columnwidth]{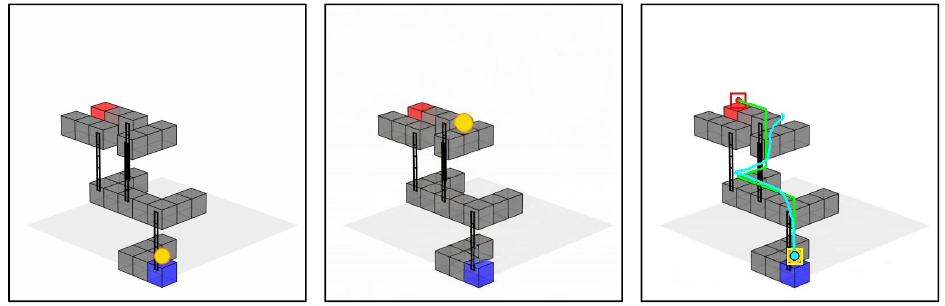} 
\caption{\textbf{Failure case: incorrect reasoning leads to unreachable path.} Although the model attempts to reach the target, its predicted trajectory (in \textcolor{BetterBlue}{blue}) passes through an infeasible region, resulting in failure to arrive at the goal (\textcolor{BetterRed}{red}). The ground-truth trajectory is shown in \textcolor{BetterGreen}{green}.}
\label{fig:fail_case1}
\end{figure}

\begin{figure}[h!]
\centering
\includegraphics[width=1\columnwidth]{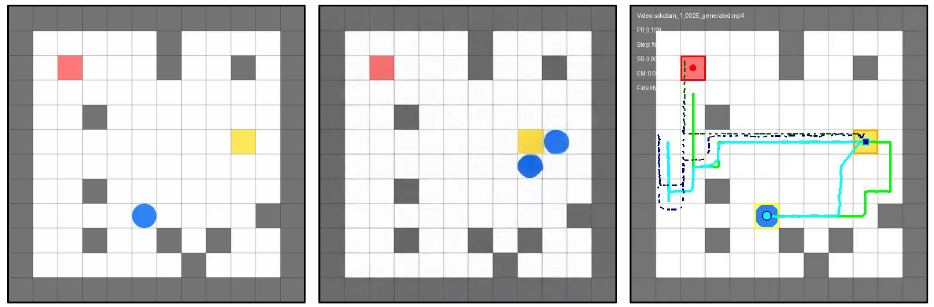} 
\caption{\textbf{Failure case: duplicated object during reasoning.} 
    During inference, the reasoning object unexpectedly appears in two locations simultaneously, indicating a visual inconsistency likely caused by faulty temporal coherence. This distracts the model and leads to an invalid reasoning path (in \textcolor{BetterBlue}{blue}), which fails to reach the correct goal (\textcolor{BetterRed}{red}).}
    \label{fig:failure_case_teleport}
\label{fig:fail_case2}
\end{figure}

\begin{figure}[h!]
\centering
\includegraphics[width=1\columnwidth]{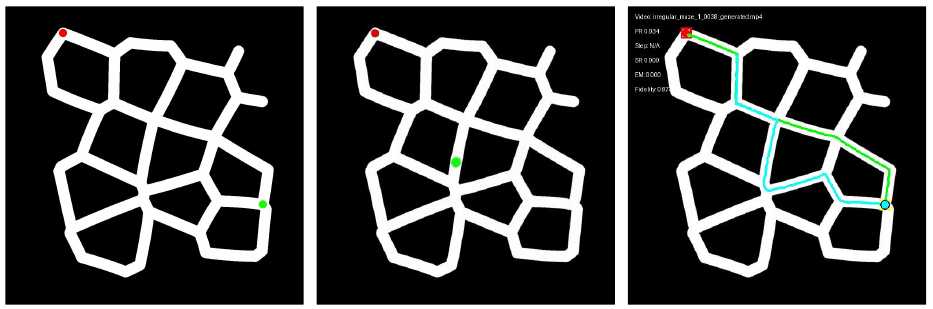} 
\caption{\textbf{Failure case: suboptimal path to target.}
    The model starts at the \textcolor{BetterGreen}{green point} and eventually reaches the \textcolor{BetterRed}{red goal}, but takes a significantly longer path than the shortest possible route. This suggests limitations in global path planning and trajectory efficiency.}
\label{fig:fail_case3}
\end{figure}

\begin{figure}[h!]
\centering
\includegraphics[width=1\columnwidth]{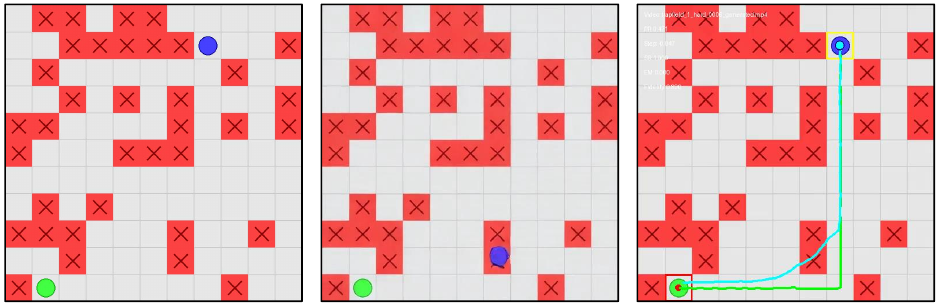} 
\caption{\textbf{Failure case: invalid path through obstacles.}
    The model starts from the \textcolor{BetterBlue}{blue point} and attempts to reach the \textcolor{BetterGreen}{green goal}, but its predicted trajectory crosses red obstacle regions, violating the environment’s physical constraints.}
\label{fig:fail_case4}
\end{figure}

\begin{figure}[h!]
\centering
\includegraphics[width=1\columnwidth]{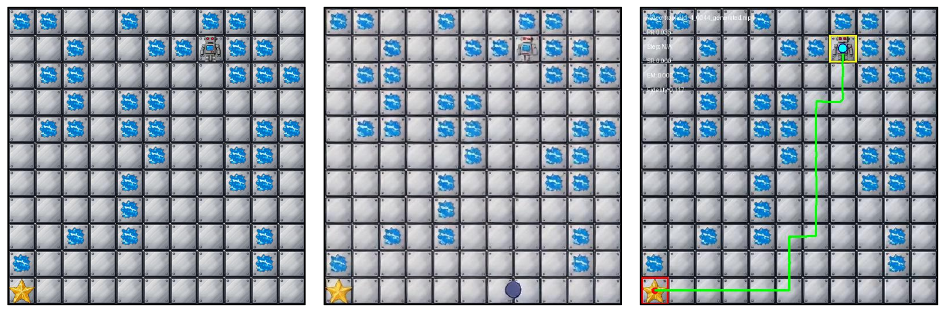} 
\caption{\textbf{Failure case: spurious object motion and goal inactivity.}
    During reasoning, the model introduces a moving object unrelated to the task, while the actual target object remains static. This reflects a failure to correctly ground the reasoning process on the intended object and leads to an incorrect trajectory.}
\label{fig:fail_case5}
\end{figure}

\begin{figure}[h!]
\centering
\includegraphics[width=1\columnwidth]{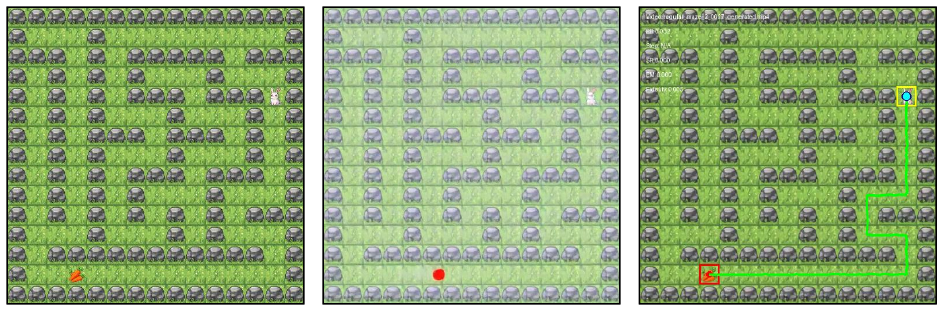} 
\caption{\textbf{Failure case: target object replaced by incorrect appearance.}
    The target object is initially a carrot, but during reasoning it is incorrectly replaced by a red dot. This severe visual inconsistency indicates the model's failure to maintain object identity, undermining its ability to reason about the correct target.}
\label{fig:fail_case6}
\end{figure}

\begin{figure}[h!]
\centering
\includegraphics[width=1\columnwidth]{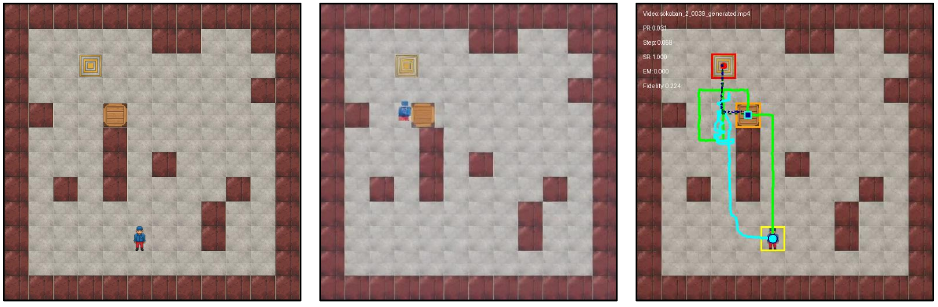} 
\caption{\textbf{Failure case: incorrect understanding of object interaction.}
    The model misinterprets the intended interaction between the character and the box. Instead of pushing the box toward the target area, the model attempts to approach the goal by pulling the box, which violates the correct physical reasoning for this task.}
\label{fig:fail_case7}
\end{figure}

\begin{figure}[h!]
\centering
\includegraphics[width=1\columnwidth]{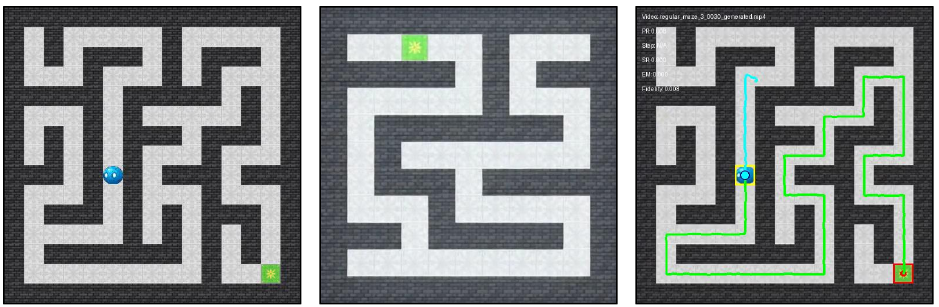} 
\caption{\textbf{Failure case: moving object disappears mid-sequence.}
    The model begins reasoning from the \textcolor{BetterBlue}{blue object} and generates a few initial frames correctly. However, the moving object then vanishes unexpectedly, leaving only the path trace without a visible agent completing the trajectory.}
\label{fig:fail_case8}
\end{figure}

\begin{figure}[h!]
\centering
\includegraphics[width=1\columnwidth]{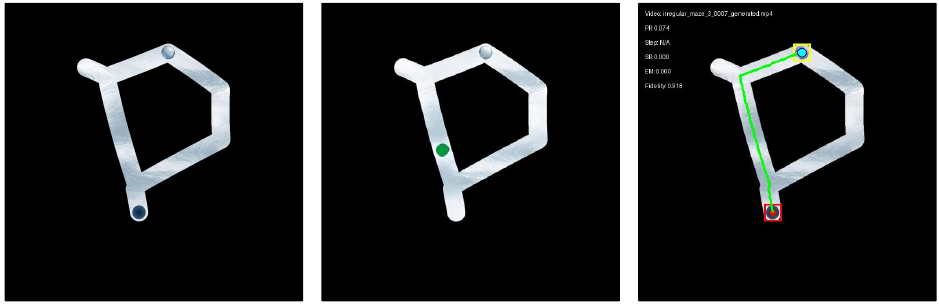} 
\caption{\textbf{Failure case: reversed motion direction and object deformation.}
    The task requires the puck to move toward the goal hole, but the model instead makes the hole move toward the puck. Moreover, the goal object is mistakenly deformed into a green sphere, indicating a failure in both physical role understanding and object appearance consistency.}
\label{fig:fail_case9}
\end{figure}

\begin{figure}[h!]
\centering
\includegraphics[width=1\columnwidth]{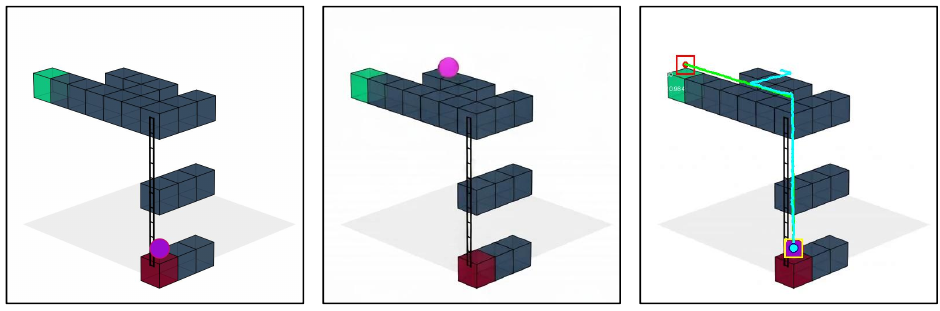} 
\caption{\textbf{Failure case: incorrect goal identification.}
    The model starts correctly from the purple ball but mistakenly identifies the wrong goal position at the top-left green region. It generates a valid path—yet toward the incorrect endpoint—resulting in a complete failure in target localization.}
\label{fig:fail_case10}
\end{figure}

\begin{figure}[h!]
\centering
\includegraphics[width=1\columnwidth]{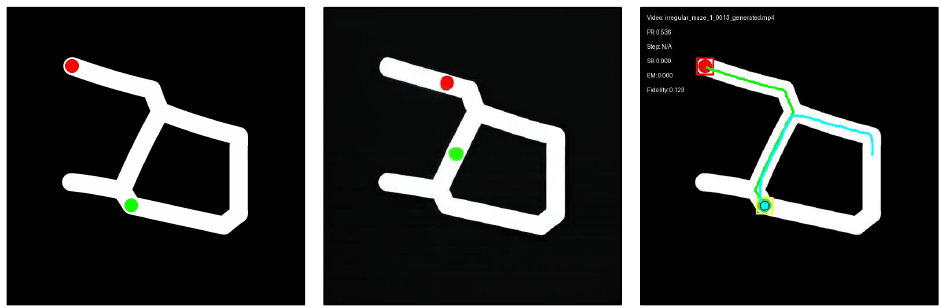} 
\caption{\textbf{Failure case: misunderstanding of reasoning target roles.}
    The model misinterprets the intended interaction: instead of moving the \textcolor{BetterGreen}{green ball} toward the \textcolor{BetterRed}{red target}, it moves both objects simultaneously. This reveals a failure in understanding asymmetric goal-directed behavior and role assignment.}
\label{fig:fail_case11}
\end{figure}

These visualization conventions are applied consistently in all subsequent failure case illustrations (see Figures~\ref{fig:fail_case1}--\ref{fig:fail_case11}).

\section{Prompt Template}
As shown in Tables, we design unified prompt templates for all five maze types (Base, Irregular, Trapfield, 3D, and Sokoban), each provided in two variants tailored for video models and VLMs. The video model prompts are presented in \textcolor{gray}{gray}, reflecting their focus on motion-generation constraints, while the VLM prompts are shown in \textcolor{blue!60}{light blue}, highlighting their emphasis on trajectory interpretation and rule compliance.

Each maze type follows a consistent prompt structure composed of a system-level prompt and task-specific instructions. For video models, the system prompt specifies the maze layout, agent behavior rules, and rendering constraints, followed by an execution prompt that defines: (1) the valid movement space; (2) expected trajectory behavior; and (3) visual requirements for temporal and spatial consistency across frames. For VLMs, the system prompt defines the evaluation context, and the task prompt formulates: (1) the extracted agent trajectory; (2) rule-checking dimensions; and (3) the expected reasoning format for judging motion validity.

Across the five maze families, these templates provide video models with precise generative instructions, while giving VLMs structured reasoning prompts for consistent evaluation.

\clearpage

\begin{table*}[t] 
\centering
\begin{tcolorbox}[
    colback=promptgray,
    colframe=promptgray,
    arc=3mm,
    boxrule=0pt,
    left=4mm,
    right=4mm,
    top=3mm,
    bottom=3mm,
    width=\textwidth
]

\textbf{Regular Maze}\\
Create a 2D animation based on the provided image of a maze. The red circle slides smoothly along the white path, stopping perfectly on the green square. The red circle never slides or crosses into the blue areas of the maze. The camera is a static, top-down view showing the entire maze.

\vspace{2mm}
\textbf{Maze:}
\begin{itemize}[leftmargin=6mm]
    \item The maze paths are white, the walls are blue.
    \item The red circle moves to the goal position, represented by a green square.
    \item The red circle slides smoothly along the white path.
    \item The red circle never slides or crosses into the blue areas of the maze.
    \item The red circle stops perfectly on the green square.
\end{itemize}

\textbf{Scene:}
\begin{itemize}[leftmargin=6mm]
    \item No change in scene composition.
    \item No change in the layout of the maze.
    \item The red circle travels along the white path without speeding up or slowing down.
\end{itemize}

\textbf{Camera:}
\begin{itemize}[leftmargin=6mm]
    \item Static camera.
    \item No zoom.
    \item No pan.
    \item No glitches, noise, or artifacts.
\end{itemize}
\end{tcolorbox}

\begin{tcolorbox}[
    colback=promptgray,
    colframe=promptgray,
    arc=3mm,
    boxrule=0pt,
    left=4mm,
    right=4mm,
    top=3mm,
    bottom=3mm,
    width=\textwidth
]

\textbf{Irregular Maze}\\
Create a 2D animation based on the provided image of a maze. The green circle slides smoothly along the white path, stopping perfectly on the red circle. The green circle never slides or crosses into the black areas of the maze. The camera is a static, top-down view showing the entire maze.

\vspace{2mm}
\textbf{Maze:}
\begin{itemize}[leftmargin=6mm]
    \item The maze paths are white, the walls are black.
    \item The green circle moves to the goal position, represented by a red circle.
    \item The green circle slides smoothly along the white path.
    \item The red circle never slides or crosses into the black areas of the maze.
    \item The green circle stops perfectly on the red circle.
\end{itemize}

\textbf{Scene:}
\begin{itemize}[leftmargin=6mm]
    \item No change in scene composition.
    \item No change in the layout of the maze.
    \item The green circle travels along the white path without speeding up or slowing down.
\end{itemize}

\textbf{Camera:}
\begin{itemize}[leftmargin=6mm]
    \item Static camera.
    \item No zoom.
    \item No pan.
    \item No glitches, noise, or artifacts.
\end{itemize}
\end{tcolorbox}

\end{table*}

\begin{table*}[t] 
\centering
\begin{tcolorbox}[
    colback=promptgray,
    colframe=promptgray,
    arc=3mm,
    boxrule=0pt,
    left=4mm,
    right=4mm,
    top=3mm,
    bottom=3mm,
    width=\textwidth
]

\textbf{3D  Maze}\\
Create a 3D animation based on the provided image of a cube maze.A yellow ball slides smoothly along the gray cube pathway, climbs up the vertical ladders step by step, and finally stops perfectly on the red cube at the top.The yellow ball never touches or passes through the blue cube or any non-gray areas of the maze.The camera remains static in an isometric, top-down angle showing the entire structure.

\vspace{2mm}
\textbf{Maze:}
\begin{itemize}[leftmargin=6mm]
    \item The maze consists of stacked transparent gray cubes forming a 3D pathway.
    \item The red cube represents the goal position. The blue cube marks the starting platform where the yellow ball begins.
    \item The yellow ball moves upward along the gray path, climbing vertically via the ladders.
    \item The ball slides smoothly without sudden changes in direction or speed.
    \item The ball stops exactly on top of the red cube at the end.
\end{itemize}

\textbf{Scene:}
\begin{itemize}[leftmargin=6mm]
    \item No structural or color changes during animation.
    \item The maze layout and cube arrangement remain unchanged.
    \item The yellow ball moves continuously at a constant speed along the 3D path.
\end{itemize}

\textbf{Camera:}
\begin{itemize}[leftmargin=6mm]
    \item Static camera.
    \item No zoom.
    \item No pan.
    \item No glitches, noise, or artifacts.
\end{itemize}
\end{tcolorbox}

\begin{tcolorbox}[
    colback=promptgray,
    colframe=promptgray,
    arc=3mm,
    boxrule=0pt,
    left=4mm,
    right=4mm,
    top=3mm,
    bottom=3mm,
    width=\textwidth
]

\textbf{Trapfield}\\
Create a 2D animation based on the provided image of a maze. The blue circle slides smoothly along the gray path, stopping perfectly on the green circle. The blue circle never slides into or crosses the red areas with crosses (trap areas). The camera is a static, top-down view showing the entire maze.

\vspace{2mm}
\textbf{Maze:}
\begin{itemize}[leftmargin=6mm]
    \item The maze paths are gray, and the trap areas are red with crosses.
    \item The blue circle moves to the goal position, represented by the green circle.
    \item The blue circle slides smoothly along the gray path.
    \item The blue circle never slides into or crosses the red areas with crosses of the maze.
    \item The blue circle stops perfectly on the green circle.
\end{itemize}

\textbf{Scene:}
\begin{itemize}[leftmargin=6mm]
    \item No change in scene composition.
    \item No change in the layout of the maze.
    \item The blue circle travels along the gray path without speeding up or slowing down.
\end{itemize}

\textbf{Camera:}
\begin{itemize}[leftmargin=6mm]
    \item Static camera.
    \item No zoom.
    \item No pan.
    \item No glitches, noise, or artifacts.
\end{itemize}
\end{tcolorbox}

\end{table*}

\begin{table*}[t] 
\centering
\begin{tcolorbox}[
    colback=promptgray,
    colframe=promptgray,
    arc=3mm,
    boxrule=0pt,
    left=4mm,
    right=4mm,
    top=3mm,
    bottom=3mm,
    width=\textwidth
]

\textbf{Sokoban}\\
Create a 2D animation based on the provided image of a grid puzzle. The blue ball moves into position behind the yellow square and smoothly pushes it toward the red goal square. The yellow square only slides when pushed from behind by the blue ball and moves in a straight line along the white floor tiles. When the direction of the yellow square’s movement needs to change, the blue ball must reposition itself to a new side of the yellow square.The square never crosses or overlaps any gray walls.

\vspace{2mm}
\textbf{Maze:}
\begin{itemize}[leftmargin=6mm]
    \item The floor area is white, and the walls are gray.
    \item The yellow square can only move when pushed by the blue ball from behind.
    \item The blue ball cannot pull the square or move through walls.
    \item The yellow square slides smoothly in one direction until it reaches the red goal square.
    \item The animation stops perfectly when the yellow square aligns with the red goal square.
\end{itemize}

\textbf{Scene:}
\begin{itemize}[leftmargin=6mm]
    \item No change in scene composition.
    \item No change in the layout of the maze.
    \item The movement is smooth, with no speed variation.
\end{itemize}

\textbf{Camera:}
\begin{itemize}[leftmargin=6mm]
    \item Static camera.
    \item No zoom.
    \item No pan.
    \item No glitches, noise, or artifacts.
\end{itemize}
\end{tcolorbox}

\end{table*}

\begin{table*}[t] 
\centering
\begin{tcolorbox}[
    colback=blue!60,
    colframe=blue!60,
    arc=3mm,
    boxrule=0pt,
    left=4mm,
    right=4mm,
    top=3mm,
    bottom=3mm,
    width=\textwidth
]

\textbf{Sokoban}\\
You are given an image of a grid-based Sokoban puzzle.\\
Gray tiles represent walls and cannot be crossed.\\
White tiles represent open floor tiles that can be moved through.\\
The blue ball represents the player or agent.\\
The yellow square represents the box that needs to be pushed.\\
The red square represents the goal destination for the box.\\

\textbf{Task:}\\
Infer the complete movement sequence required for the blue ball to push the yellow square onto the red goal square.\\
The blue ball moves in four directions: up, down, left, right.\\
When the blue ball moves into a box, it automatically pushes the box if there is space behind it.\\
The box and the blue ball cannot cross or overlap any gray walls.\\
Diagonal movement is not allowed, and the camera remains fixed from a top-down view.\\

\textbf{Output Format:}\\
Return the entire movement sequence as a JSON array of directional actions, where each element is one of ``up'', ``down'', ``left'', or ``right''.\\
Do not include any explanations or additional text.\\

\textbf{Example of expected output:}\\
\{
  "actions": ["right", "right", "down", "left", "down"]
\}

\end{tcolorbox}
\end{table*}

\begin{table*}[t] 
\centering

\begin{tcolorbox}[
    colback=blue!60,
    colframe=blue!60,
    arc=3mm,
    boxrule=0pt,
    left=4mm,
    right=4mm,
    top=3mm,
    bottom=3mm,
    width=\textwidth
]
\textbf{Regular Maze}\\
You are given an image of a grid-based maze.\\
Black tiles represent walls and cannot be crossed.\\
White tiles represent open paths that can be moved through.\\
The green circle represents the starting point of the path.\\
The red circle represents the goal or destination.\\

\textbf{Task:}\\
Infer the shortest valid path from the green starting point to the red goal circle.\\
Movement can only occur between adjacent open tiles — up, down, left, or right.\\
Diagonal movement is not allowed, and the path must not cross or touch any black walls.\\

\textbf{Output Format:}\\
Return the entire movement sequence of the green circle as a JSON array of directions,\\
where each element is one of ``up'', ``down'', ``left'', or ``right''.\\
Do not include any explanations or additional text.\\

\textbf{Example of expected output:}\\
\{
  "path": ["up", "up", "left", "down", "right", "right"]
\}
\end{tcolorbox}

\vspace{2mm}

\begin{tcolorbox}[
    colback=blue!60,
    colframe=blue!60,
    arc=3mm,
    boxrule=0pt,
    left=4mm,
    right=4mm,
    top=3mm,
    bottom=3mm,
    width=\textwidth
]
\textbf{Irregular Maze}\\
You are given an image of a pathfinding puzzle.\\
The image shows a network of curved paths connecting various waypoints.\\
Each waypoint (intersection or junction) is labeled with a letter or letter combination\\
(A, B, C, ..., Z, AA, AB, etc.).\\
The green circle represents the starting point.\\
The red circle represents the goal or destination.\\

\textbf{Task:}\\
Find the shortest valid path from the green starting point to the red goal.\\
The path must follow the visible roads/paths in the image.\\
You can only move along the connected paths shown in the image.\\

\textbf{Output Format:}\\
You MUST return a JSON object with a ``path'' field containing an array of waypoint labels.\\
The array should start with the label closest to the starting point\\
and end with the label closest to the goal.\\
Do not include any explanations or additional text.\\
Important: The ``path'' field MUST be an array of strings, not a single string.\\

\textbf{Example of expected output:}\\
\{
  "path": ["A", "B", "C", "D", "E"]
\}
\end{tcolorbox}

\end{table*}

\begin{table*}[t] 
\centering

\begin{tcolorbox}[
    colback=blue!60,
    colframe=blue!60,
    arc=3mm,
    boxrule=0pt,
    left=4mm,
    right=4mm,
    top=3mm,
    bottom=3mm,
    width=\textwidth
]

\textbf{Maze3D}\\
You are given an image of a 3D maze composed of gray cubes that represent walkable platforms suspended in space.\\
Each cube represents a solid tile that the ball can stand on or move across.\\
The yellow sphere represents the starting point.\\
The blue cubes represent the initial platform where the ball begins.\\
The red cube represents the goal or destination.\\

\textbf{Task:}\\
Infer the shortest valid 3D path for the yellow sphere to move from its starting position to the red goal cube.\\

\textbf{Movement Rules:}\\
Horizontal movements (forward\_left, forward\_right, backward\_left, backward\_right): each move spans 2 grid units horizontally.\\
Vertical movements (up, down): each move spans 3 grid units vertically via a ladder; a ladder must be present at the starting position.\\
The sphere cannot move through empty space or overlap any cube structure.\\
All movements must follow valid cube surfaces and ladder connections.\\

The six valid directions of movement are:\\
``forward\_left'' -- move diagonally forward and to the left (2 units) within the same layer.\\
``forward\_right'' -- move diagonally forward and to the right (2 units) within the same layer.\\
``backward\_left'' -- move diagonally backward and to the left (2 units) within the same layer.\\
``backward\_right'' -- move diagonally backward and to the right (2 units) within the same layer.\\
``up'' -- move vertically upward (3 units) via a ladder.\\
``down'' -- move vertically downward (3 units) via a ladder.\\

\textbf{Output Format:}\\
Return the full sequence of movement directions as a JSON array, where each step is one of the six valid directions.\\
Do not include any explanations, reasoning, or extra text.\\

\textbf{Example of expected output:}\\
\{
  "path": ["up", "forward\_right", "forward\_left", "up", "forward\_right"]
\}

\end{tcolorbox}

\vspace{2mm}

\begin{tcolorbox}[
    colback=blue!60,
    colframe=blue!60,
    arc=3mm,
    boxrule=0pt,
    left=4mm,
    right=4mm,
    top=3mm,
    bottom=3mm,
    width=\textwidth
]

\textbf{Trapfield}\\
You are given an image of a grid-based maze.\\
Red tiles marked with an ``X'' represent trap zones that must be avoided.\\
White tiles represent open paths that can be moved through.\\
The blue circle represents the starting point of the path.\\
The green circle represents the goal or destination.\\

\textbf{Task:}\\
Infer the shortest valid path for the blue circle to reach the green circle.\\
Movement can only occur between adjacent open tiles -- up, down, left, or right.\\
Diagonal movement is not allowed.\\
The path must not cross or touch any red trap tiles.\\

\textbf{Output Format:}\\
Return the full movement sequence of the blue circle as a JSON array of directions,\\
where each element is one of ``up'', ``down'', ``left'', or ``right''.\\
Do not include any explanations, reasoning, or extra text.\\

\textbf{Example of expected output:}\\
\{
  "path": ["left", "left", "down", "down"]
\}

\end{tcolorbox}

\end{table*}

\end{document}